\documentclass[10pt,twocolumn,letterpaper]{article}

\usepackage[pagenumbers]{cvpr}

\usepackage{microtype}
\usepackage{multirow}
\usepackage{tabularx}
\usepackage{flafter}
\usepackage{booktabs}
\usepackage{array}
\usepackage{amsmath}
\usepackage{algorithm}
\usepackage{algorithmic}
\newlength{\tilew}

\newcommand{\sref}[1]{\S\ref{#1}}
\newcommand{\eref}[1]{Eq.~\ref{#1}}
\newcommand{\tref}[1]{Table~\ref{#1}}
\newcommand{\fref}[1]{Figure~\ref{#1}}

\definecolor{cvprblue}{rgb}{0.21,0.49,0.74}
\usepackage[pagebackref,breaklinks,colorlinks,allcolors=cvprblue]{hyperref}
\AtEndPreamble{
  \crefname{figure}{Figure}{Figures}
  \Crefname{figure}{Figure}{Figures}
  \crefformat{section}{\S#2#1#3}
  \Crefformat{section}{\S#2#1#3}
  \crefformat{subsection}{\S#2#1#3}
  \Crefformat{subsection}{\S#2#1#3}
  \crefmultiformat{section}{\S#2#1#3}{ and~#2#1#3}{, #2#1#3}{, and~#2#1#3}
  \Crefmultiformat{section}{\S#2#1#3}{ and~#2#1#3}{, #2#1#3}{, and~#2#1#3}
  \crefformat{appendix}{\S#2#1#3}
  \Crefformat{appendix}{\S#2#1#3}
  \crefformat{subappendix}{\S#2#1#3}
  \Crefformat{subappendix}{\S#2#1#3}
  \crefmultiformat{appendix}{\S#2#1#3}{ and~#2#1#3}{, #2#1#3}{, and~#2#1#3}
  \Crefmultiformat{appendix}{\S#2#1#3}{ and~#2#1#3}{, #2#1#3}{, and~#2#1#3}
  \crefname{table}{Table}{Tables}
  \Crefname{table}{Table}{Tables}
  \creflabelformat{equation}{#2#1#3}
}

\def\confName{CVPR}
\def\confYear{2027}

\title{StyleComposer: Training-Free Multi-Reference Style Composition}

\author{Sanghyeok Lee \qquad Jihye Kang \qquad Namhyuk Ahn\\
Inha University}

\begin{document}
\twocolumn[{%
\renewcommand\twocolumn[1][]{#1}%
\maketitle
\vspace{-2.5em}
\begin{center}
\captionsetup{type=figure,skip=4pt}
\includegraphics[width=\textwidth]{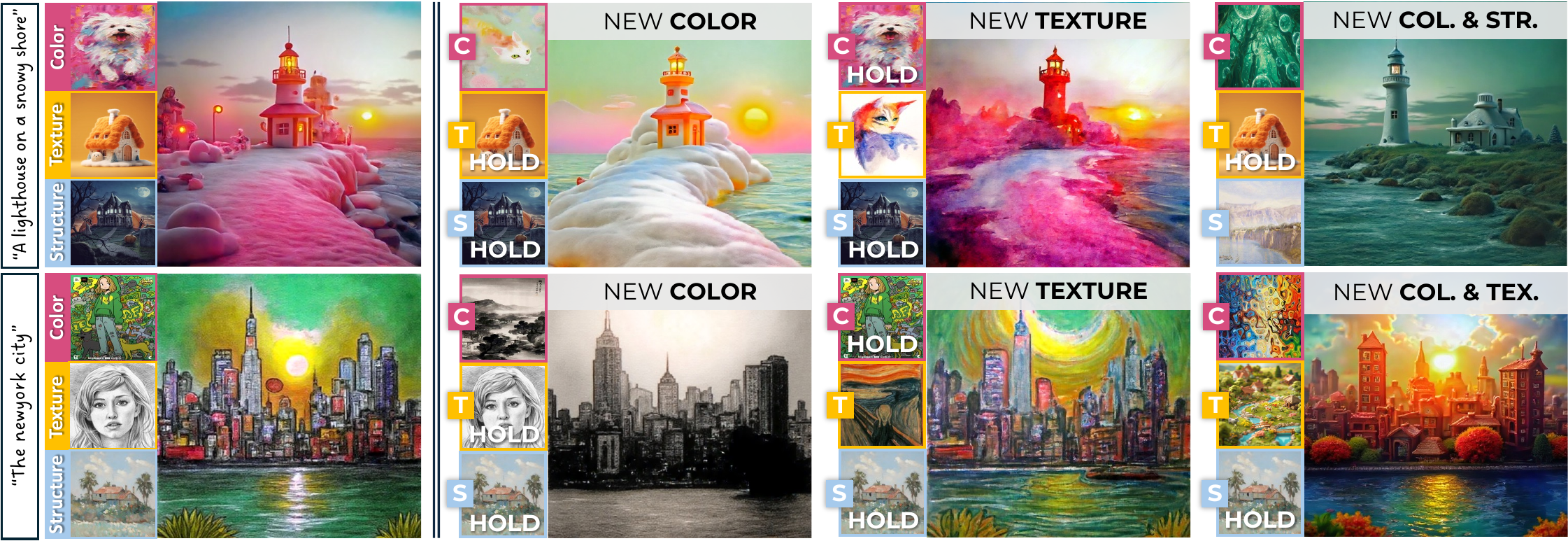}
\captionof{figure}{\textbf{Style attribute composition under reference replacement.}
Each row starts from one set of color (C), texture (T), and structure (S) references, and each later column replaces the references named in its header while \textsc{Hold} marks the references kept fixed.
The outputs follow the replaced attributes and largely keep those of the held references.}
\label{fig:teaser}
\end{center}
\vspace{0.9em}
}]

\begin{abstract}
The color, texture, and structure of an image can come from different references, but reference-guided generators transfer each reference as a single style signal.
We find that in a diffusion transformer, every pathway that brings in a reference transfers multiple attributes, yet each attribute has one representation in which it is most directly controllable.
Guided by this finding, we design StyleComposer to route color through the chromatic coordinates of the latent, texture through the attention keys and values, and structure through the queries, and to limit what else each route admits from its reference.
Without training or inversion, StyleComposer follows the references and the prompt together more closely than stylization methods and multi-image generators, and it sets the strength of each attribute separately.
\end{abstract}

\section{Introduction}

\emph{The style of a painting is not monolithic.}
An artist may take color from one image, texture from another, and structure from a photograph, each at a chosen strength.
Art pedagogy breaks a picture down into elements~\cite{ocvirk2012art,itten1975design}, and we focus on three of them, \emph{color} (the palette and tonal distribution), \emph{texture} (the brushwork, grain, and medium), and \emph{structure} (the arrangement and scale of regions in the picture plane).
Reference-guided stylization methods render a prompt in the style of a reference image through per-reference tuning, image adapters, or shared attention features~\cite{ahn2024dreamstyler,wang2024instantstyle,hertz2024style,zhang2026alignedgen}.
They reach high fidelity, and they transfer the three elements of a reference together as one style signal.
Choosing a style therefore comes down to finding a single image that already holds the desired palette, medium, and arrangement.

Methods that go beyond a single style signal offer this freedom only in part.
Learned methods fix how a reference splits into attributes at training time~\cite{frenkel2024implicit,wu2024fiva}, and training-free methods separate at most two roles, typically color and style~\cite{qin2025free,liu2026disentangling}.
Slider methods set the strength of a concept, but each slider is trained for one concept and has no reference input~\cite{gandikota2024concept}.
Multi-image generators~\cite{wu2025omnigen2,flux2klein2026,qwenimage2026} take several images, each with a written role, and seem to offer a direct solution.
Given a watercolor of a cat as the color reference, however, two of them paint the cat itself into a requested lighthouse scene, although the reference should supply only its palette.
Naming the attribute in the instruction does not stop the other content of a reference from entering the image.

Taking each attribute from its own reference at its own strength lets an artist revise a style one attribute at a time.
When the palette misses the intent, the artist replaces only the color reference, and the medium and the arrangement largely stay as they were (\cref{fig:teaser}).
A strength for each attribute sets how closely the output follows each reference, so the artist can keep only a hint of a palette or push a medium further.
A small pool of references per attribute combines into many styles, and none of them has to exist as a single image.
We call this task \textbf{style attribute composition}, which generates an image from a prompt and separate color, texture, and structure references, with a strength for each attribute.
Art courses teach color, texture, and composition as separate subjects~\cite{itten1975design,ocvirk2012art}, so each reference stands for one decision that an artist already makes on its own.

The difficulty of the task lies in the references themselves.
Each reference holds all three attributes while it is designated for only one, so the generation must take that attribute and leave the other two behind, together with the objects of the reference.
We call the transfer of an attribute other than the designated one \textbf{attribute leakage}, as when the color reference also brings in its brushwork.
With leakage, replacing a reference or raising a strength also changes attributes that it is not meant to change.
Composition succeeds only when the designated attributes and the prompt are satisfied together without leakage, so a method that reproduces one reference faithfully while it loses another has not composed anything.
Preventing leakage requires knowing where each attribute enters the generator, so that each reference enters only through the pathway of its attribute.

We locate these entry points by probing a diffusion transformer (DiT)~\cite{peebles2023scalable}.
Every pathway that brings in a reference transfers more than one attribute, yet each attribute has one representation in which it is most directly controllable.
Texture follows the token-wise pairing of the attention keys and values, color follows the global statistics of the values and occupies three chromatic coordinates of the VAE latent, and the queries hold the layout of a reference together with its content.
On the basis of this finding, StyleComposer routes each attribute through its own representation, restricts each route to what its attribute needs, and assigns each route a strength of its own, all without training or inversion.
It follows the references and the prompt together more closely than stylization methods, dual-reference methods, and multi-image generators.
Replacing one reference moves the output toward the new reference on that attribute, and raising one strength moves the output toward its reference mainly on that attribute.
Our contributions are as follows.
\begin{itemize}
\item We introduce the task of style attribute composition with a strength for each attribute and evaluate it by joint satisfaction, reference replacement, and strength response.
\item We probe DiT pathways and find that each attribute has one representation where it is most directly controllable.
\item We propose StyleComposer, a training-free method that routes each attribute through that representation and outperforms all compared methods in joint satisfaction.
\end{itemize}

\section{Related Work}

\noindent\textbf{Reference-guided stylization.}
Diffusion-based stylization methods produce reference styles through per-reference tuning~\cite{ahn2024dreamstyler,sohn2023styledrop}, image adapters~\cite{ye2023ip,wang2024instantstyle}, shared attention features~\cite{hertz2024style,zhang2026alignedgen,chung2024style}, or large-scale stylization data~\cite{wang2025omnistyle}, extending even to video~\cite{ye2025stylemaster}. Despite their differences, most treat each reference as a single style signal, transferring color, texture, and spatial structure together and limiting separate control over the source and strength of each attribute.

\smallskip
\noindent\textbf{Attribute-level and multi-reference control.}
Learned methods fix the separation of style attributes at training time, using block-specific LoRA~\cite{frenkel2024implicit}, adapters trained on attribute annotations~\cite{wu2024fiva}, or predicted style LoRA~\cite{duan2026compressing}.
Training-free dual-reference methods take a color reference and a style reference, separated by arithmetic on CLIP image embeddings in SADis~\cite{qin2025free}, by frequency bands in Dual-FDM~\cite{liu2026disentangling}, or by latent color transforms in \citet{agarwal2024training}.
Multi-image generators~\cite{wu2025omnigen2,flux2klein2026,qwenimage2026} condition one generation on several images and assign their roles through a written instruction.
StyleComposer assigns three roles to three separate images and gives each role a strength of its own to adjust.

\smallskip
\noindent\textbf{Internal control of DiTs.}
Studies of DiTs assign roles to the queries, keys, and values of attention~\cite{yin2025consistedit}, to the layers that depend on rotary positions~\cite{wei2025freeflux}, and to a linear color subspace of the VAE latent~\cite{pach2026latent}.
Methods that build on these roles transfer the whole appearance of a reference through shared attention~\cite{hertz2024style,zhang2026alignedgen}, limit copying from a reference by scaling rotary frequencies~\cite{mikaeili2026untwisting}, or steer structure with query features~\cite{lin2025freecontrol,alaluf2024cross}.
We measure which attributes each pathway admits when a reference enters it, assign each attribute to the pathway that controls it most directly, and restrict what else that pathway admits from the reference.

\section{Method}
\label{sec:method}

\begin{figure*}[t]
\centering
\includegraphics[width=\textwidth]{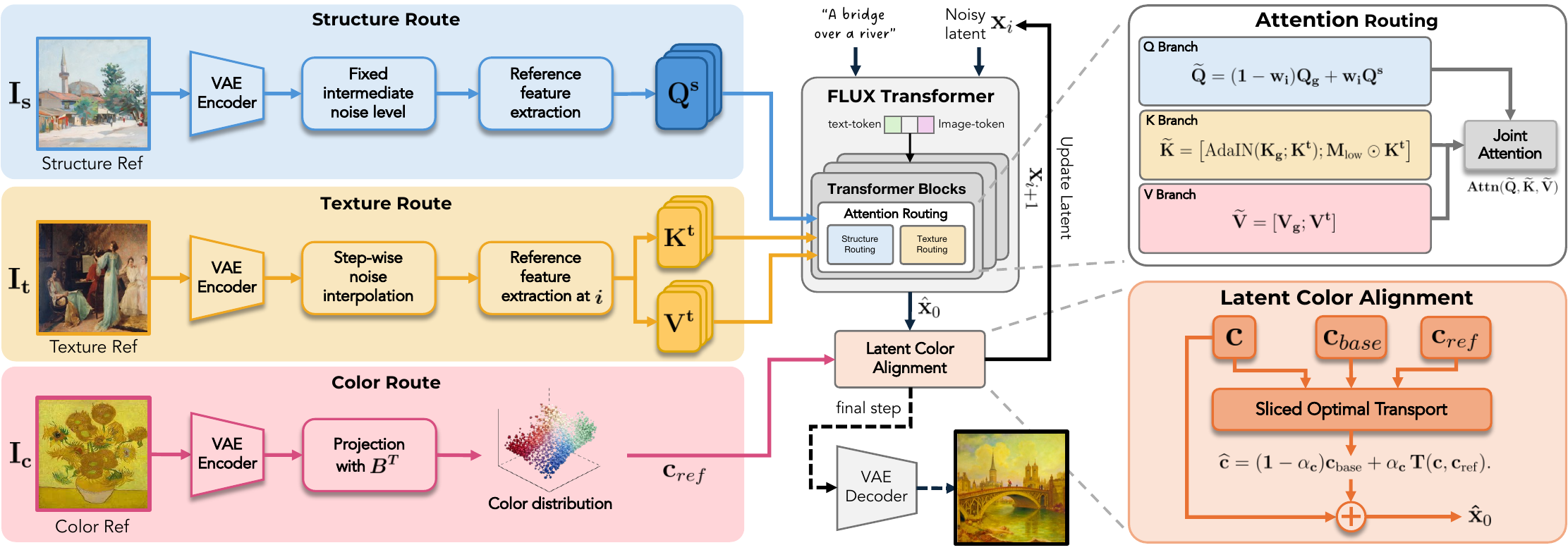}
\caption{\textbf{Overview of StyleComposer.} The structure route records the queries $Q^s$ of the structure reference once and blends them into the generation queries. The texture route brings the texture reference to the current noise level and appends its keys $K^t$ and values $V^t$ to the joint attention. The color route projects the clean estimate $\hat{x}_0$ and the color reference onto three chromatic coordinates and transports the distribution of the generation to that of the color reference.}
\label{fig:pipeline}
\end{figure*}

StyleComposer generates an image from a text prompt $p$, a color reference $I_c$, a texture reference $I_t$, and a structure reference $I_s$.
It builds on a diffusion transformer (DiT)~\cite{peebles2023scalable} that processes text and image tokens together in joint attention,
\begin{equation}
\mathrm{Attn}(Q,K,V)=\mathrm{Softmax}\!\left(\frac{QK^\top}{\sqrt{d}}\right)V.
\end{equation}
At sampling step $i$, the transformer predicts a velocity $v_i$, and $\hat{x}_0=x_i-\sigma_i v_i$ is its clean estimate of the final image.
To control each attribute separately without attribute leakage, we first locate the representation in the DiT that holds each attribute (\cref{sec:diagnosis}) and then route each reference through the representation of its attribute (\S\ref{sec:color}--\ref{sec:structure}).

\subsection{Diagnosis and Principle}
\label{sec:diagnosis}

\noindent\textbf{Probe setup.}
In a DiT, a reference reaches the generation most directly through joint attention, so we first measure what its keys and values transfer.
We bring the reference latent to the current noise level, record its image-token keys and values, and append them to those of the generation in every block and step.
For each attribute, the movement toward the reference is the fraction of the distance between the output without injection and the reference that the injection closes.
The movement is zero when the injection leaves the attribute unchanged and one when the output matches the reference in that attribute.

\begin{figure}[t]
\centering
\includegraphics[width=\linewidth]{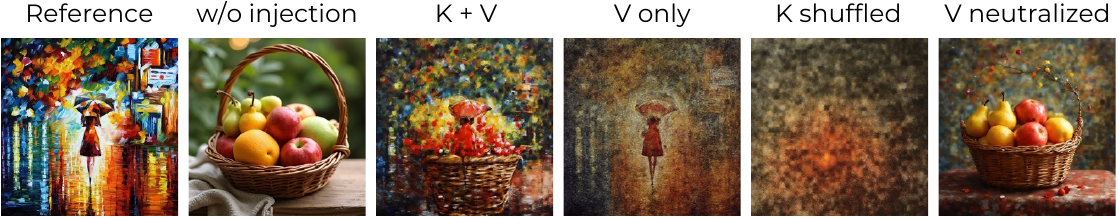}
\caption{\textbf{What reference keys and values transfer.} The prompt asks for a fruit basket, and the reference is a painting. Appending the raw keys and values of the reference (K + V) brings its palette, its brushwork, and its figure under the umbrella into the output, and appending its values alone (V only) replaces the scene with the reference. With shuffled reference keys (K shuffled), the pairing of keys and values breaks, and a warm palette remains without the brushwork. With the value statistics matched to the generation (V neutralized), natural colors return and a painterly grain remains.}
\label{fig:channels}
\end{figure}

\smallskip
\noindent\textbf{Keys and values.}
Appending the raw keys and values moves all three attributes toward the reference and also brings in the reference objects (\cref{fig:channels}).
Two controlled changes locate the attributes within this pathway.
Shuffling the reference keys and keeping the values in place breaks the token-wise pairing of keys and values, and the texture movement nearly vanishes while most of the color movement stays.
Matching the channel-wise mean and variance of the reference values to those of the generation removes the color movement and keeps about half of the texture movement.
Texture therefore follows the token-wise pairing of keys and values, and color follows the global statistics of the values.

\smallskip
\noindent\textbf{Queries.}
Queries determine where each generated token attends, and cross-image attention transfers appearance by combining the queries of a structure image with the keys and values of an appearance image~\cite{alaluf2024cross}.
We record the reference queries once from the noise-free reference latent at a fixed step and substitute them for the generation queries at every step.
Full substitution moves the layout toward the reference as far as the raw keys and values do and moves color and texture less, yet it halves the prompt similarity of the output (results in Suppl.).
Queries therefore hold layout together with appearance and content, so layout control through the queries needs its own restrictions on when and where it acts.

\begin{table*}[t]
\centering
\footnotesize
\caption{\textbf{Routes of StyleComposer.} Each row gives, for one attribute, the representation through which its route brings in the reference, the mechanism the route builds on, the restriction we add to keep the other attributes of the reference out, the sampling steps in which the route acts (0 to 27), and its strength.
The restrictions are what separate each route from the mechanism it builds on.}
\label{tab:routes}
\begin{tabularx}{\textwidth}{@{}l>{\raggedright\arraybackslash}p{2.6cm}>{\raggedright\arraybackslash}X>{\raggedright\arraybackslash}X cc@{}}
\toprule
Route & Representation & Builds on & Our restriction & Steps & Strength \\
\midrule
Color & Chromatic coordinates of $\hat{x}_0$ & Latent color subspace~\cite{pach2026latent} and sliced optimal transport~\cite{rabin2011wasserstein} & Edits 3 of 16 latent dimensions after each transformer pass; strength anchored to a reference-free prediction & 16--27 & $\alpha_c$ \\
Texture & Correspondence between keys and values & Noised reference keys and values~\cite{zhang2026alignedgen} and rotary frequency scaling~\cite{mikaeili2026untwisting} & High-frequency key components attenuated; statistics aligned on keys only; attention cap in early steps & 3--27 & $s_{\mathrm{lf}}$ \\
Structure & Queries & One-step attention extraction~\cite{lin2025freecontrol} and structure queries~\cite{alaluf2024cross} & Blended with a cosine decay in early steps; last 25 single-stream blocks only & 0--11 & $T_s$ \\
\bottomrule
\end{tabularx}
\end{table*}

\begin{figure*}[t]
\centering
\includegraphics[width=\textwidth]{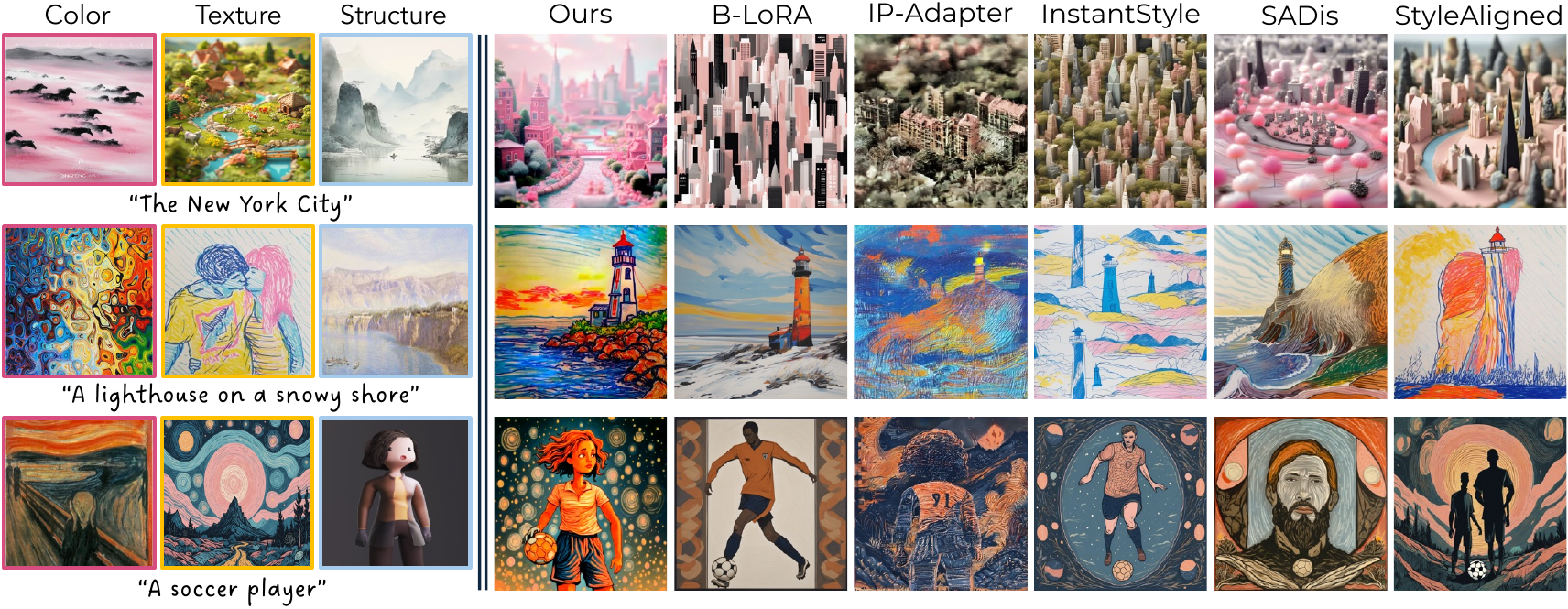}
\caption{\textbf{Qualitative comparison on three-attribute composition.}
Each row shows color, texture, and structure references with a prompt.
The baselines transfer appearance or content of the references beyond the designated attributes, as in the miniature cities of the first row, whereas ours follows each designated attribute and keeps the prompted objects.}
\label{fig:qualitative}
\end{figure*}

\smallskip
\noindent\textbf{Latent color subspace.}
Outside attention, color has a more direct handle.
The VAE latent of Flux~\cite{flux2024} contains a three-dimensional linear subspace that encodes color~\cite{pach2026latent}.
We encode 512 solid-color images and average each latent over positions, and the first three principal components of these averages form the orthonormal basis $B\in\mathbb{R}^{16\times3}$ with mean $\mu$ and explain 97\% of their variance.
Hue, saturation, and value edits of five scenes place 93\% of the change in the latent distribution within these three dimensions, whereas an isotropic change would place 19\% there.
Color thus has three directly movable coordinates per latent position.

\smallskip
\noindent\textbf{Principle.}
These measurements lead to one principle.
Every pathway that brings in a reference transfers more than one attribute, yet each attribute has a representation in which it is most directly addressable, namely the latent \emph{chromatic coordinates} for color, the \emph{token-wise correspondence between keys and values} for texture, and the \emph{queries} for structure.
StyleComposer brings each reference into the generation only through the representation of its attribute, and we call this path, together with the restrictions that keep the other attributes out, the \emph{route} of that attribute.
The \emph{texture route} drops the positional detail of the reference keys and aligns statistics on the keys alone, the \emph{structure route} acts only in the early steps, and the \emph{color route} edits three coordinates after every transformer pass and thereby overwrites the palette that the other two routes bring in (\cref{tab:routes}).

\subsection{Color Route}
\label{sec:color}

The color route makes the palette of the generation follow the color reference.
At each color step, the route projects the clean estimate onto the basis, $c_u=B^\top(\hat{x}_{0,u}-\mu)$ at each latent position $u$, and the latent $z^c_0$ of the color reference gives $c^{\mathrm{ref}}_u=B^\top(z^c_{0,u}-\mu)$ in the same way.
A reference supplies a distribution of colors over its positions, so the route transports the distribution of $c_u$ to that of $c^{\mathrm{ref}}_u$.
Sliced optimal transport~\cite{rabin2011wasserstein} performs this transport by matching quantiles along random rotations of the three axes, which respects the correlation among the coordinates, and we denote the transported coordinates by $\mathcal{T}(c_u,c^{\mathrm{ref}}_u)$.

The color strength $\alpha_c$ interpolates between the transported coordinates and the coordinates $c^{\mathrm{base}}_u$ of a second clean estimate that the model predicts from the same noisy latent with every reference route disabled,
\begin{equation}
\bar{c}_u=(1-\alpha_c)c^\mathrm{base}_{u}+\alpha_c\mathcal{T}(c_u,c^{\mathrm{ref}}_u).
\label{eq:alpha}
\end{equation}
Only the chromatic displacement returns to the latent,
\begin{equation}
\hat{x}_{0,u}\leftarrow\hat{x}_{0,u}+B(\bar{c}_u-c_u).
\label{eq:lift}
\end{equation}
Because $B$ has orthonormal columns, the other 13 latent dimensions of $\hat{x}_0$ keep exactly the values that the transformer predicted.
At $\alpha_c=0$, the route keeps the palette that the model predicts from the current state without any reference route, so lowering $\alpha_c$ also removes the palette that the attention routes add during the color steps.
The route acts in the late steps (\cref{tab:routes}), and after each update the sampler converts the estimate back to a velocity and takes its usual Euler step toward the next noise level.

\begin{table*}[t]
\centering
\footnotesize
\setlength{\tabcolsep}{5pt}
\caption{\textbf{Quantitative comparison} on the benchmark.
\textit{Composition} and \textit{Selectivity} are defined in \cref{sec:setup}, Human is the share of user-study votes, and the other columns are distances to each reference and to the prompt.
gCSD and gDINO are grayscale CSD and DINO, and IP-Adapter adds a depth ControlNet when a structure reference is given.
Bold and underline mark the best and second-best values.}
\label{tab:main}
\begin{tabular*}{\textwidth}{@{\extracolsep{\fill}}lcccccccccc@{}}
\toprule
Method & \multicolumn{2}{c}{Style Composition} & Human & \multicolumn{2}{c}{Color}
& \multicolumn{2}{c}{Texture} & \multicolumn{2}{c}{Structure} & Prompt \\
\cmidrule(lr){2-3}\cmidrule(lr){4-4}\cmidrule(lr){5-6}\cmidrule(lr){7-8}\cmidrule(lr){9-10}\cmidrule(lr){11-11}
& \textit{Compo.}$\uparrow$ & \textit{Select.}$\uparrow$ & Pref.\,(\%)$\uparrow$
& MS-SWD$\downarrow$ & C-Hist$\downarrow$
& gCSD$\downarrow$ & 1$-$CLIP-I$\downarrow$
& depth$\downarrow$ & gDINO$\downarrow$
& 1$-$CLIP-T$\downarrow$ \\
\midrule
Flux~\cite{flux2024} & 0.347 & 0.434 & 6.3 & 9.51  & 0.651 & 0.590 & 0.375 & \underline{0.158} & 0.257 & \underline{0.762} \\
SDXL~\cite{podell2024sdxl} & 0.407 & 0.430 & 5.3 & 7.83  & 0.587 & 0.608 & 0.374 & 0.193 & 0.245 & 0.786 \\
\midrule
InstantStyle~\cite{wang2024instantstyle} & 0.243 & 0.324 & 10.3 & 14.29 & 0.801 & 0.431 & 0.326 & 0.181 & 0.269 & 0.799 \\
StyleAligned~\cite{hertz2024style} & 0.274 & 0.440 & 9.5 & 9.56  & 0.669 & \textbf{0.368} & \textbf{0.272} & 0.278 & 0.258 & 0.820 \\
IP-Adapter~\cite{ye2023ip} & 0.303 & 0.370 & 13.7 & 10.98 & 0.707 & 0.487 & 0.319 & 0.256 & \textbf{0.229} & 0.813 \\
SADis~\cite{qin2025free} & \underline{0.432} & \underline{0.446} & \underline{20.0} & \underline{7.46} & \underline{0.571} & \underline{0.403} & 0.319 & 0.229 & 0.251 & 0.802 \\
B-LoRA~\cite{frenkel2024implicit} & 0.301 & 0.391 & 1.5 & 8.12 & 0.627 & 0.672 & 0.398 & 0.188 & 0.260 & \textbf{0.761} \\
\midrule
\textbf{StyleComposer}
& \textbf{0.621} & \textbf{0.745} & \textbf{33.3} & \textbf{5.32}
& \textbf{0.443} & 0.439 & \underline{0.315}
& \textbf{0.088} & \underline{0.232} & 0.785 \\
\bottomrule
\end{tabular*}
\end{table*}

\subsection{Texture Route}
\label{sec:texture}

The texture route transfers the rendering of the texture reference, meaning its medium, brushwork, and grain, through the correspondence between keys and values.
The route brings the reference latent to the current noise level, $x^t_i=(1-\sigma_i)z^t_0+\sigma_i\epsilon^t$ with noise $\epsilon^t$ fixed per generation, so it needs no inversion~\cite{zhang2026alignedgen}.
A forward pass of $x^t_i$ gives the image-token keys $K^t$ and values $V^t$ of the current step, and every generated query attends to them together with the keys and values of the generation, $\mathrm{Attn}(Q_g,[K_g;\widetilde{K}^t],[V_g;V^t])$.
The route runs in every attention block and starts after the first few steps, so that the prompt and the structure route set the arrangement first.
Raw keys and values also bring in the layout and the objects of the texture reference (\cref{sec:diagnosis}), so the route adds the following three restrictions.

First, the route attenuates the high-frequency positional components of the reference keys.
High-frequency rotary components let a generated token find the reference token at the same position, which copies objects and layout from a shared reference~\cite{mikaeili2026untwisting}.
We set the high-frequency end of the rotary frequency scaling of \citet{mikaeili2026untwisting} to zero and scale the $j$-th of the $n$ frequency pairs of each spatial axis, ordered from the highest to the lowest frequency, by
\begin{equation}
m_j=s_{\mathrm{lf}}\big(j/(n-1)\big)^{2},\quad \widetilde{K}^t=\mathrm{diag}(m)\,\mathrm{RoPE}(K^t),
\label{eq:lowk}
\end{equation}
which raises the lowest frequency to the texture strength $s_{\mathrm{lf}}$.
This suppresses fine positional matching, so a generated token matches reference tokens by their appearance over broad regions.
Because $s_{\mathrm{lf}}$ multiplies every positional component, it scales the attention logits of the reference tokens and with them how strongly generated tokens match the reference.

Second, the route aligns statistics on the keys alone.
Appended reference keys come from a different image, so their statistics differ from those of the generation keys and they compete poorly for attention.
StyleAligned~\cite{hertz2024style} aligns the channel-wise mean and variance of both queries and keys to the reference, and we align only the image-token keys of the generation, separately for each head and channel,
\begin{equation}
K_g^{\mathrm{img}}\leftarrow\mathrm{AdaIN}\big(K_g^{\mathrm{img}};K^t\big).
\label{eq:adain}
\end{equation}
The queries stay unaligned, because aligning them as well pulls them toward the reference and imports its layout.
The values stay untouched, because their global statistics hold color (\cref{sec:diagnosis}) and aligning them would add the palette of the texture reference to the output.

Third, the route caps the share of attention that goes to the reference tokens in the early steps.
For a generated query $i$, the route computes the attention outputs $O^g_i$ and $O^t_i$ over the generated keys and over the reference keys separately, and it mixes them with the weight $a_i$ that the joint softmax would give the reference tokens, capped at $\tau_i$,
\begin{equation}
O_i=(1-\tilde{a}_i)O^g_i+\tilde{a}_iO^t_i,\qquad \tilde{a}_i=\min(a_i,\tau_i),
\label{eq:cap}
\end{equation}
where $\tau_i$ is tightest in the early steps, looser in the middle steps, and lifted in the late steps.
Early reference tokens would otherwise override the arrangement set by the prompt and the structure route, so the cap reserves the early steps for them and lets texture accumulate later.

\begin{figure*}[t]
\centering
\includegraphics[width=\textwidth]{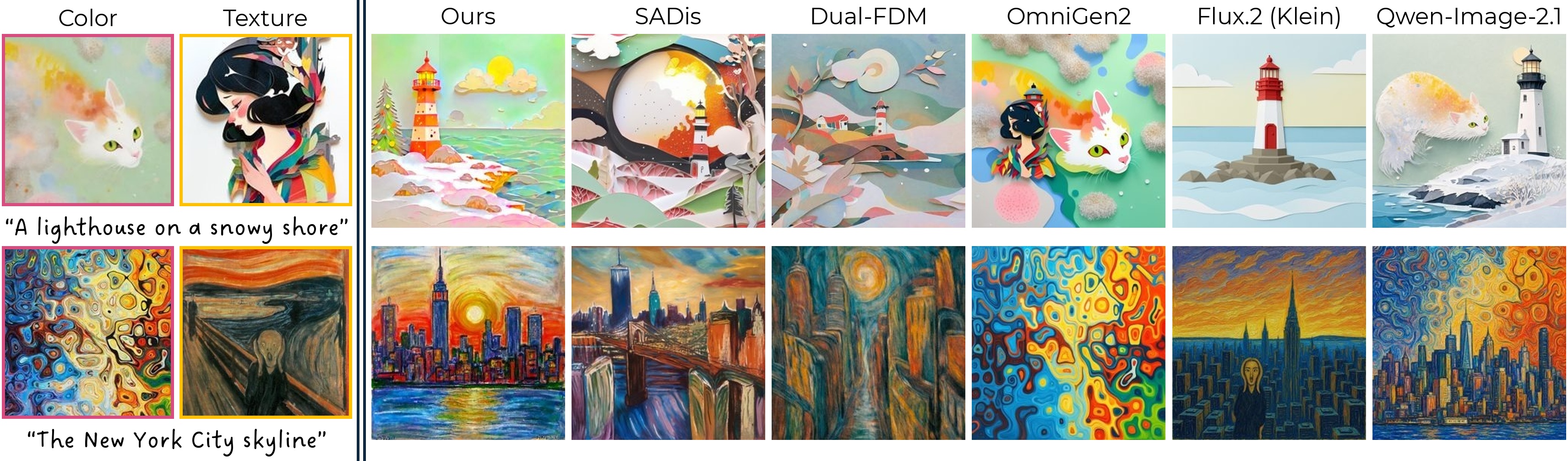}
\caption{\textbf{Qualitative comparison on two-reference composition.}
Each row shows a color reference and a texture reference with a prompt.
The multi-image generators and SADis copy objects and shapes of the references into the scene, such as the cat, the paper-cut girl, the screaming figure, and the bridge railing.
Ours follows the palette and the medium and keeps the prompted subject.}
\label{fig:two_reference}
\end{figure*}

\begin{figure*}[t]
\centering
\includegraphics[width=\textwidth]{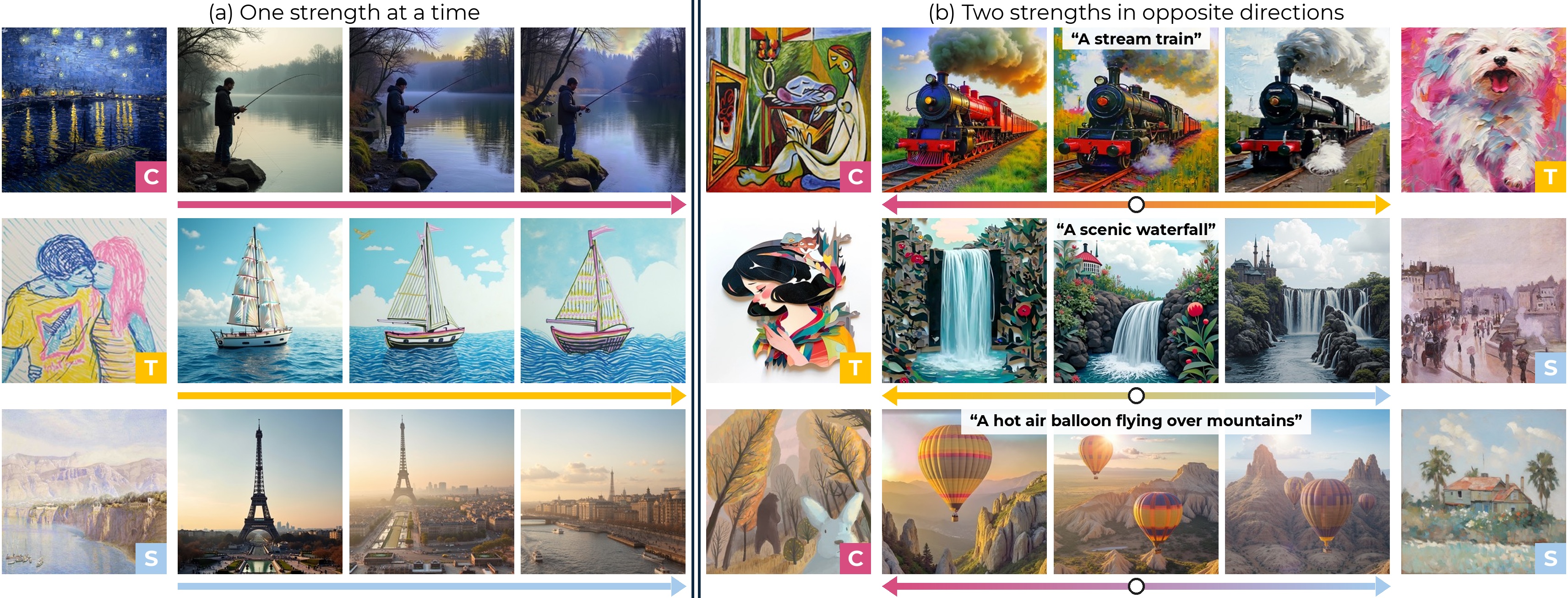}
\caption{\textbf{Per-attribute strength.}
In each row, the references, the prompt, and the seed are fixed, and C, T, and S mark the color, texture, and structure references.
(a) The strength of one attribute increases along the arrow.
(b) The strength of the left reference decreases and that of the right reference increases along the arrow from left to right.}
\label{fig:control}
\end{figure*}

\subsection{Structure Route}
\label{sec:structure}

The structure route transfers the coarse arrangement of the structure reference through the queries.
Blending queries changes where generated tokens attend without appending reference tokens, so the values of the structure reference, which hold its palette and texture, stay out of the attention.
The route records the image-token queries $Q^s$ of the structure reference once, from a single forward pass of its noise-free latent at a fixed step~\cite{lin2025freecontrol}.
In each of the deeper single-stream blocks, the route blends this anchor into the generation queries with a weight $w_i$ that follows a cosine from one to zero at step $T_s$,
\begin{equation}
Q_g^{(\ell)}\leftarrow(1-w_i)Q_g^{(\ell)}+w_iQ^{s,(\ell)},
\label{eq:qblend}
\end{equation}
where the weight is $w_i=\frac{1}{2}[1+\cos(\pi i/T_s)]$ for $i<T_s$.
The route uses only the deeper single-stream blocks, since the earlier ones transfer the layout about as well but pull the output away from the prompt.
The cosine weight follows from the diagnosis that queries hold content and color together with layout.
It starts at one, so the anchor sets the arrangement in the first steps, and it reaches zero at step $T_s$, after which the generation uses its own queries and the details follow the prompt.
The anchor is noise-free because a noised reference shows little of its layout in the first steps.

\subsection{Per-Attribute Strength}
\label{sec:strength}

Each route exposes a single strength parameter, and each parameter enters only its own route.
The color strength $\alpha_c$ moves the palette between the reference-free prediction and the transported reference palette (\cref{eq:alpha}), the texture strength $s_{\mathrm{lf}}$ scales the attention logits of the reference tokens (\cref{eq:lowk}), and the structure strength $T_s$ sets how many steps the query anchor lasts (\cref{eq:qblend}).
All benchmark cases share one setting of the three strengths, and leaving out a reference turns its route off.
The routes add one forward pass of the texture reference per texture step and one reference-free forward pass per color step, so a generation takes 1.8 times as long as prompt-only sampling with Flux.

\section{Experiments}
\label{sec:experiments}

\begin{table}[t]
\centering
\footnotesize
\setlength{\tabcolsep}{4pt}
\caption{\textbf{Two-reference composition} on 100 cases.
Every method receives the same color reference, texture reference, and prompt, and MS-SWD, gCSD, and 1$-$CLIP-T are the distances to these three.
\textit{Composition} is normalized within this comparison, so its values are not comparable with those of \cref{tab:main}.
Bold and underline mark the best and second-best values.}
\label{tab:two_reference}
\begin{tabular*}{\linewidth}{@{\extracolsep{\fill}}lcccc@{}}
\toprule
Method & \textit{Compo.}$\uparrow$ & MS-SWD$\downarrow$ & gCSD$\downarrow$ & 1$-$CLIP-T$\downarrow$ \\
\midrule
SADis~\cite{qin2025free} & \underline{0.671} & \underline{7.45} & \textbf{0.339} & 0.793 \\
Dual-FDM~\cite{liu2026disentangling} & 0.584 & 9.23 & \underline{0.368} & 0.797 \\
\midrule
OmniGen2~\cite{wu2025omnigen2} & 0.248 & 14.07 & 0.377 & 0.821 \\
Flux.2 (Klein)~\cite{flux2klein2026} & 0.291 & 14.56 & 0.445 & 0.796 \\
Qwen-Image-2.1~\cite{qwenimage2026} & 0.478 & 10.08 & 0.479 & \underline{0.779} \\
\midrule
StyleComposer & \textbf{0.741} & \textbf{5.46} & 0.470 & \textbf{0.758} \\
\bottomrule
\end{tabular*}
\end{table}

\subsection{Experimental Setup}
\label{sec:setup}
We evaluate 610 fixed combinations of a prompt with one, two, or three references drawn from pools of five WikiArt, StyleDrop, and DreamStyler images per attribute~\cite{artgan2018,sohn2023styledrop,ahn2024dreamstyler}, and 250 of them use all three references.
We compare against SADis~\cite{qin2025free}, InstantStyle~\cite{wang2024instantstyle}, StyleAligned~\cite{hertz2024style}, IP-Adapter Flux~\cite{ye2023ip}, B-LoRA~\cite{frenkel2024implicit}, and the prompt-only Flux and SDXL.
Each method receives the references that its image inputs accept and the remaining attributes as text, and StyleComposer adds the medium of the texture reference to the prompt.
We measure color with MS-SWD~\cite{he2024multiscale} and a color histogram, texture with grayscale CSD~\cite{somepalli2024measuring} and CLIP-I~\cite{radford2021learning}, structure with depth layout~\cite{yang2024depth} and grayscale DINO~\cite{caron2021emerging}, and prompt alignment with CLIP-T.
\textit{Composition} is the harmonic mean of the normalized similarities to the prompt and to every requested reference, so an output that satisfies one reference while losing another scores near zero, and \textit{Selectivity} is the share of the change toward the references that falls on the designated attributes (details in Suppl.).
Among DiTs, we build StyleComposer on Flux~\cite{flux2024} throughout the main paper.

\subsection{Model Comparison}

In \cref{fig:qualitative,tab:main}, StyleComposer follows each designated attribute while it keeps the prompted objects, and it reaches the highest \textit{Composition} and \textit{Selectivity}.
Yet InstantStyle~\cite{wang2024instantstyle}, StyleAligned~\cite{hertz2024style}, and SADis~\cite{qin2025free} have a smaller texture distance than StyleComposer, because they reproduce the texture reference almost as it is, as the miniature city in the first row of \cref{fig:qualitative} shows.
Copying that reference pulls their outputs away from the color and structure references and from the prompt, so their color, depth, and CLIP-T distances are all larger than those of StyleComposer and their \textit{Composition} stays low.
In a user study, 30 participants viewed 20 three-attribute cases and chose for each case the output that best matched the three references while following the prompt, and they chose ours in 33.3\% of the 600 judgments, against 20.0\% for the runner-up.

\begin{table}[t]
\centering
\footnotesize
\setlength{\tabcolsep}{3pt}
\caption{\textbf{Reference replacement}, averaged over 40 pairs of outputs per row.
Each row replaces one reference and keeps everything else fixed.
Response is $R$ of \cref{eq:response} for the replaced attribute, Drift is the change of each held attribute between the paired outputs, and $\Delta$CLIP-T is the change in prompt similarity.}
\label{tab:replacement}
\begin{tabular*}{\linewidth}{@{\extracolsep{\fill}}lccccc@{}}
\toprule
& & \multicolumn{3}{c}{Drift} & \\
\cmidrule(lr){3-5}
Replaced & Response$\uparrow$ & Color & Texture & Structure & $\Delta$CLIP-T \\
\midrule
Color & 17.47 & -- & 0.171 & 0.013 & $+$0.0045 \\
Texture & 0.194 & 2.56 & -- & 0.052 & $+$0.0005 \\
Structure & 0.038 & 0.67 & 0.143 & -- & $-$0.0011 \\
\bottomrule
\end{tabular*}
\end{table}

The advantage holds without the three-reference setting or the medium descriptor.
StyleComposer has the highest \textit{Composition} for every combination of requested attributes, from a single reference to all three, and without the medium descriptor added to its prompt it still reaches a \textit{Composition} of 0.562 and a \textit{Selectivity} of 0.701, above every compared method (results in Suppl.).
Combining strong single-attribute controls does not close the gap either.
Histogram matching on the outputs of IP-Adapter gives the lowest color distance, and a Canny ControlNet added to SADis gives the lowest depth distance, yet both combinations stay at a \textit{Composition} of 0.330 or below (results in Suppl.).

\smallskip
\noindent\textbf{Two-reference composition.}\label{sec:two_reference}
Two further groups of methods take a color reference and a texture reference without a structure reference.
SADis~\cite{qin2025free} and Dual-FDM~\cite{liu2026disentangling} keep a color reference and a style reference apart within one model.
The multi-image generators OmniGen2~\cite{wu2025omnigen2}, Flux.2 (Klein)~\cite{flux2klein2026}, and Qwen-Image-2.1~\cite{qwenimage2026} receive the instruction to use the first image only for the color palette and the second only for texture and artistic style.
We form 100 cases from our five color references, five texture references, and four prompts, and every method receives the same prompt followed by the medium descriptor of the texture reference.

StyleComposer reaches the highest \textit{Composition} and is the closest of all methods to the color reference and to the prompt (\cref{tab:two_reference}).
SADis and Dual-FDM are closer to the texture reference, and \cref{fig:two_reference} shows that they also reproduce shapes of the references, such as the round paper cutout around the lighthouse.
The multi-image generators have the lowest \textit{Composition} even though the instruction names the role of each image, because they paint objects of the references into the scene, such as the cat of the color reference and the screaming figure of the oil painting.
A written role therefore does not keep the content of a reference out of the generation, and the restrictions of each route in StyleComposer provide this separation.

\begin{table}[t]
\centering
\footnotesize
\setlength{\tabcolsep}{4pt}
\caption{\textbf{Route ablation} on 54 three-reference cases.
For each route, the row gives the distance of its attribute to its reference with and without the route.
A positive $\Delta$ means that the output is farther from the reference without the route.}
\label{tab:route_ablation}
\begin{tabular*}{\linewidth}{@{\extracolsep{\fill}}llccc@{}}
\toprule
Removed route & Metric & Full & w/o route & $\Delta$ \\
\midrule
Color alignment & MS-SWD$\downarrow$
& \textbf{5.887} & 15.572 & $+9.685$ \\
Texture K/V & gCSD$\downarrow$
& \textbf{0.442} & 0.608 & $+0.166$ \\
Structure Q & depth$\downarrow$
& \textbf{0.119} & 0.139 & $+0.020$ \\
\bottomrule
\end{tabular*}
\end{table}

\subsection{Model Analysis}
\noindent\textbf{Reference replacement.}\label{sec:replacement}
Reference replacement tests directly whether each attribute comes from its own reference.
We replace one reference at a time, as in \cref{fig:teaser}, and keep the other references, the prompt, the seed, and the strengths fixed.
If an attribute comes from its reference, the output moves toward the new reference on that attribute and changes little on the others.
We measure this movement for the replaced attribute with the response
\begin{equation}
R=\tfrac{1}{2}\big[(d_{AB}-d_{BB})+(d_{BA}-d_{AA})\big],
\label{eq:response}
\end{equation}
where $y_A$ and $y_B$ are the outputs before and after replacing the reference $r_A$ with $r_B$, and $d_{ij}=d(y_i,r_j)$ is the distance of the replaced attribute.
The first term measures how much closer the output gets to the new reference and the second how much farther it gets from the old one, so $R>0$ means that the output follows the replacement.
We replace each reference in 40 pairs of outputs from 20 combinations and two seeds, using prompts that do not name the medium of the texture reference (details in Suppl.).

$R$ is positive for all three attributes (\cref{tab:replacement}), so each output follows the reference that replaced its attribute.
\Cref{fig:teaser} shows the same behavior.
A pastel color reference turns the pink lighthouse scene pastel while its clay-like rendering and its layout remain, and a new texture reference turns the skyline into the swirling strokes of the oil painting while its palette stays.
Replacing two references at once moves both of their attributes, such as the green palette and the rocky shoreline that new color and structure references give the lighthouse scene.
The palette of the output depends mainly on the color reference.
The output colors move 17.5 toward a new color reference, whereas a new texture or structure reference changes them by only 2.6 and 0.7.
Prompt alignment also holds, since CLIP-T changes by less than 0.005.

\smallskip
\begin{table}[t]
\centering
\footnotesize
\setlength{\tabcolsep}{5pt}
\caption{\textbf{Design ablation} on the same 54 cases.
Each row changes one design choice of the full model.
Every change lowers \textit{Composition}, and bold and underline mark the best and second-best values.}
\label{tab:ablation}
\begin{tabular*}{\linewidth}{@{\extracolsep{\fill}}lcc@{}}
\toprule
Variant & \textit{Composition}$\uparrow$ & \textit{Selectivity}$\uparrow$ \\
\midrule
Single K/V route & 0.175 & 0.371 \\
Overlapped route windows & 0.150 & \textbf{0.611} \\
Full-band texture K & 0.061 & 0.357 \\
Q/K AdaIN & \underline{0.351} & 0.508 \\
W/o texture-attention cap & 0.325 & 0.515 \\
\midrule
Full (ours) & \textbf{0.583} & \underline{0.585} \\
\bottomrule
\end{tabular*}
\end{table}

\noindent\textbf{Style sliders.}
The three route strengths act as style sliders.
In \cref{fig:control}a, the references, the prompt, and the seed stay fixed while raising one slider moves the output toward its reference on that attribute.
The palette goes from a hint of the reference colors to the full palette, the rendering goes from a plain image to the full medium, and the layout shifts from the prompted scene to the arrangement of the structure reference.
On 20 reference--prompt combinations with two seeds each, raising one strength from its lowest setting to the benchmark setting lowers the distance of its own attribute by 62\% for color, 26\% for texture, and 48\% for structure.
Over the same range, the other two distances change by at most 16\%, and CLIP-T drops by at most 0.017 (more results in Suppl.).
Moving two sliders in opposite directions shifts the emphasis between attributes (\cref{fig:control}b), for example from a palette-driven to a texture-driven rendition, and each setting costs a single generation.

\smallskip
\noindent\textbf{Ablation study.}
The ablations test the reasons that \cref{sec:method} gives for each route and each restriction.
Removing a route degrades its own attribute, and neither the prompt nor the other two routes recover it (\cref{tab:route_ablation}).
Without the color route the color distance grows 2.6 times, without the texture route the texture distance rises by 38\% even though the prompt still names the medium of the texture reference, and without the structure route the depth distance rises by 17\%.
Describing an attribute in the prompt does not replace its route either, because the color distance then doubles and the depth distance grows by 58\% (results in Suppl.).

Removing a restriction lowers \textit{Composition}, because the references then override the arrangement or the content that the prompt sets (\cref{tab:ablation}).
Restoring the full band of the texture keys lets the texture reference impose its own layout and causes the largest drop, aligning the queries together with the keys imports the same layout, and removing the attention cap lets early reference tokens override the arrangement.
Replacing the three routes with one shared key--value route lowers both \textit{Composition} and \textit{Selectivity}, as the principle of \cref{sec:diagnosis} predicts.
Overlapping the route windows is the one variant with a higher \textit{Selectivity}, since the always-on routes keep the transfer selective, yet they suppress the prompt-driven content and leave \textit{Composition} near its lowest value.

\raggedbottom
\section{Conclusion}
We present StyleComposer, a training-free method for composing color, texture, and structure from separate references, each at its own strength.
Our analysis shows that attribute selectivity depends on the representation, motivating a color route through the chromatic coordinates of the latent, a texture route through the attention keys and values, and a structure route through the queries, each restricted to what its attribute needs.
StyleComposer achieves the highest \textit{Composition} and \textit{Selectivity} among the compared methods, and exposes one style slider per attribute that sets its strength.
A replaced reference moves the output on its attribute, and each slider moves mainly its own attribute, so an artist can revise a style one attribute at a time.

\vspace{\dimexpr\smallskipamount+4.7pt\relax}
\noindent\textbf{Limitations.}
Art practice distinguishes elements finer than our three attributes, such as line quality and surface texture within texture or shape and spatial organization within structure~\citep{ocvirk2012art,itten1975design}.
Two references can already share one route, as when a marker drawing and a yarn-like surface are mixed through the texture route (\cref{fig:supp_limitations}), and giving each finer element a route of its own is a direction for future work.

\section*{LLM Usage Statement}
We used Claude Code with Claude Opus 5.5 to edit and refine the author-written manuscript, implementation code, and the analysis of experimental results.
We also obtained reviewer-style feedback on the full manuscript from Claude Code and Codex.
The research ideas, hypotheses, claims, and conclusions are the authors' own, and LLM assistance on them was limited to editing and refining their presentation.
The authors take full responsibility for all submitted content, regardless of any LLM or AI-tool use.

{
    \small
    \bibliographystyle{ieeenat_fullname}
    \bibliography{main}
}

\clearpage
\setcounter{page}{1}
\flushbottom
\appendix
\renewcommand{\thefigure}{A\arabic{figure}}\renewcommand{\thetable}{A\arabic{table}}
\renewcommand{\dblfloatpagefraction}{0.25}\renewcommand{\floatpagefraction}{0.25}

\section{Benchmark and Metrics}
\label{sec:supp_benchmark}

\subsection{Benchmark Construction}
\label{sec:supp_benchmark_construction}

\begin{figure}[t]
\centering
\includegraphics[width=\linewidth]{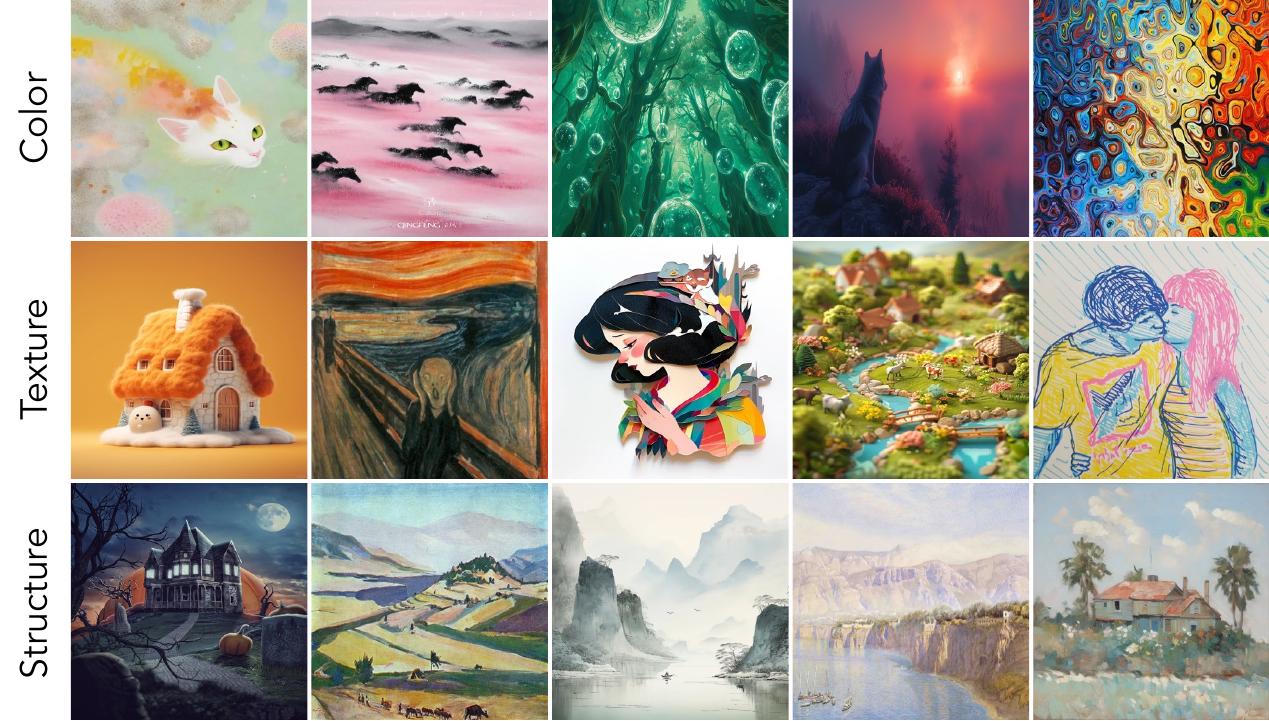}
\caption{\textbf{Reference image pool used in the composition benchmark.}
The benchmark draws color, texture, and structure from separate pools of five images. Each row holds one role, and its references are indexed 1--5 from left to right.}
\label{fig:supp_reference_pool}
\end{figure}

Our benchmark uses disjoint pools of five color, five texture, and five
structure references (\cref{fig:supp_reference_pool}). The color pool spans pastel, monochromatic, cool, gradient, and high-contrast palettes, the texture pool spans felt, oil pastel, paper-cut, miniature diorama, and marker renderings, and the structure pool contains distinct coarse arrangements of the major regions and depth layers. 

We evaluate all three single-attribute requests, all three attribute pairs,
and the three-reference setting. For each single attribute, five references
are combined with four content prompts, giving 20 cases per attribute and 60
in total. For each attribute pair, all $5\times5$ reference pairs are
combined with four prompts, giving 100 cases per pair and 300 in total. The
three-reference tier contains all $5^3$ reference triplets under two prompts,
giving 250 cases. The complete benchmark therefore contains
$60+300+250=610$ fixed cases. The four content prompts are
``A lighthouse on a snowy shore'',
``A sailboat on the ocean'',
``The New York City skyline'', and
``A scenic waterfall''; the first and third are used for the
three-reference tier. All outputs are generated with seed 42.

\subsection{Inputs of Compared Methods}
\label{sec:supp_input_mapping}

\begin{table*}[t]
\centering
\footnotesize
\setlength{\tabcolsep}{5pt}
\caption{\textbf{Method-specific reference mapping.}
Each method retains its native reference mechanism. For image-conditioned
baselines, requests for which none of the designated references can enter
that mechanism are left unevaluated rather than represented only by text.}
\label{tab:supp_input_mapping}
\begin{tabularx}{\textwidth}{@{}l>{\raggedright\arraybackslash}X>{\raggedright\arraybackslash}X>{\raggedright\arraybackslash}X@{}}
\toprule
Method & Reference mechanism & Text-side input & Requests (cases) \\
\midrule
StyleComposer
& C, T, and S through their dedicated routes
& T medium descriptor when T is active
& All seven nonempty combinations (610) \\
SADis
& C and T through its designated image inputs
& S layout description
& All except S only (590) \\
IP-Adapter Flux
& T through IP-Adapter; S through an added depth ControlNet
& C palette description
& All except C only (590) \\
InstantStyle
& T through its style-image input
& C palette and S layout descriptions
& T, C+T, T+S, C+T+S (470) \\
StyleAligned
& T through inversion and shared attention
& C palette, S layout, and its native style phrase
& T, C+T, T+S, C+T+S (470) \\
B-LoRA
& T through a per-reference style LoRA
& C palette, S layout, and its learned style token
& T, C+T, T+S, C+T+S (470) \\
Flux / SDXL
& None
& Descriptions of every requested attribute
& All seven nonempty combinations (610) \\
\bottomrule
\end{tabularx}
\end{table*}

\begin{table*}[t]
\centering
\footnotesize
\setlength{\tabcolsep}{8pt}
\caption{\textbf{Exact attribute descriptions used by text-conditioned
inputs.} Index $k$ is the $k$-th reference in each row of \cref{fig:supp_reference_pool}.}
\label{tab:supp_attribute_descriptors}
\begin{tabularx}{\textwidth}{@{}c>{\raggedright\arraybackslash\hsize=.85\hsize}X>{\raggedright\arraybackslash\hsize=.65\hsize}X>{\raggedright\arraybackslash\hsize=1.5\hsize}X@{}}
\toprule
$k$ & Color palette & Texture medium & Structure layout \\
\midrule
1 &
mint green, ivory, pale peach, powder blue, soft gray, pastel &
soft felt 3D rendering &
with a central angular block framed by branching shapes on the left,
a curved path leading up from below, and a large arc rising behind \\
2 &
blush pink, black, white, charcoal gray, muted monochrome &
rough oil pastel painting &
with curved bands sweeping across the lower frame, a central ridge
overlapping them, and a larger mass receding above \\
3 &
emerald green, teal, mint, dark green, white highlights, cool vivid &
layered paper-cut art &
with tall masses on both sides, pointed layers receding toward the
center, and a broad open channel below \\
4 &
coral pink, magenta, indigo, navy, peach, warm-cool contrast &
miniature diorama rendering &
with a layered ridge along the top, a long mass extending to the right,
a broad basin below, and a small cluster at the lower left \\
5 &
electric blue, orange, yellow, red, black, white, high saturation &
felt-tip marker drawing &
with a broad central block flanked by two tall forms, a dense band
below, and an open upper half \\
\bottomrule
\end{tabularx}
\end{table*}

The compared methods expose different reference interfaces. We keep the native reference mechanism of each method and give a short text description for a requested attribute that has no matching image input.
A purely textual request does not count as a supported reference transfer for an image-conditioned baseline, whereas Flux~\cite{flux2024} and SDXL~\cite{podell2024sdxl} are reported separately as text-only baselines.
We do not synthesize pseudo-references for unsupported slots. \tref{tab:supp_input_mapping} summarizes the protocol, where the labels C, T, and S denote the color, texture, and structure references.

For IP-Adapter Flux~\cite{ye2023ip}, we add depth ControlNet~\cite{zhang2023adding} only when structure is requested,
which makes it a stronger baseline than IP-Adapter alone. For
StyleAligned~\cite{hertz2024style}, InstantStyle~\cite{wang2024instantstyle}, and B-LoRA~\cite{frenkel2024implicit}, the single available style mechanism is assigned to the texture reference. SADis~\cite{qin2025free} uses its separate
color and texture inputs. These assignments are fixed across all cases and
are not selected per example. StyleComposer also appends the medium descriptor of the texture reference to its prompt, and \cref{sec:supp_texture_prompt} repeats the benchmark without this descriptor.

\smallskip
\noindent\textbf{Prompt construction.}\label{sec:supp_prompts}
Each benchmark case stores an original content prompt $p$, a short palette
description $d_c$, a texture-medium descriptor $d_t$, and a coarse layout
description $d_s$. Palette descriptions list the dominant colors and broad
tonal relation. Texture descriptions identify the rendering medium, and
layout descriptions name only coarse relative placement, extent, and depth
organization without naming objects from the structure reference. All
descriptors are manually checked against their references. \Cref{tab:supp_attribute_descriptors} gives the exact text, where index $k$ is the $k$-th reference in each row of \cref{fig:supp_reference_pool}.
The generic text substitutions are
\begin{align}
p_c &= p+\texttt{, in a palette of }+d_c, \nonumber\\
p_t &= p+\texttt{, in }+d_t+\texttt{ style}, \nonumber\\
p_s &= p+\texttt{, }+d_s,
\label{eq:supp_prompt_templates}
\end{align}
with only the requested fields included. Method-specific required suffixes,
including InstantStyle's
``, masterpiece, best quality, high quality'' suffix,
StyleAligned's per-reference style phrase, and B-LoRA's learned
\texttt{[v]} style token, are retained from their official implementations.

StyleComposer uses $p+\texttt{, }+d_t$ when texture is active and $p$
otherwise; it does not append $d_c$ or $d_s$. The texture phrase names only the medium and no content of the reference.
For every method, prompt alignment is measured against the original content prompt $p$, before any attribute description or method-specific suffix is appended.

\subsection{Metrics}
\label{sec:supp_metrics}

\noindent\textbf{Attribute distances.}
We measure color with MS-SWD, using the official implementation at five scales with 128 fixed projections, and with the total-variation distance between joint $8\times8\times8$ RGB histograms. Texture is measured on grayscale images by one minus CSD ViT-L style similarity and one minus CLIP ViT-L/14 image similarity. Structure uses depth-layout distance and the mean
absolute difference between grayscale DINO ViT-B/8 self-similarity matrices.
For the depth-layout distance, we resize the Depth Anything V2 Small depth maps to $128\times128$, standardize them independently, and compare them as
\begin{equation}
D_s(I,R)=1-\frac{1}{HW}\sum_{u}
\overline d_I(u)\,\overline d_R(u),
\label{eq:supp_depth_layout}
\end{equation}
which is one minus their spatial correlation. Prompt alignment is one minus
CLIP ViT-L/14 image--text similarity to the original content prompt. All
reported primitive metrics are therefore distances for which lower is
better.

\smallskip
\noindent\textbf{\textit{Composition}.}
For each primitive metric, we pool the available outputs of the eight methods
over all supported benchmark tiers, compute its 5th and 95th percentiles
$(l_m,h_m)$, and define the clipped similarity
\begin{equation}
s_m(x)=1-
\operatorname{clip}\!\left(\frac{x-l_m}{h_m-l_m},0,1\right).
\label{eq:supp_metric_normalization}
\end{equation}
Every method uses these boundaries. For an image with active attribute set $\mathcal A$, \textit{Composition}
is the harmonic mean of the similarity to the original prompt and the primary-metric similarity of every active attribute,
\begin{equation}
\mathrm{Composition}
=
\frac{|\mathcal A|+1}
{s_p^{-1}+\sum_{a\in\mathcal A}s_a^{-1}}.
\label{eq:supp_composition}
\end{equation}
The primary metrics are MS-SWD, grayscale CSD distance, and depth-layout
distance for color, texture, and structure, respectively. Scores are first
computed per image and then averaged; similarities are lower-bounded by
$10^{-6}$ when evaluating the harmonic mean. Thus, one poorly satisfied
requirement cannot be hidden by another.

\begin{figure}[t]
\centering
\includegraphics[width=\linewidth]{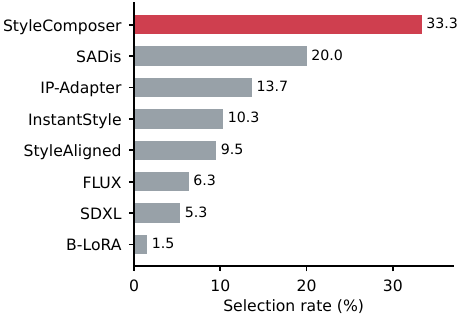}
\caption{\textbf{Complete user-preference results.}
Each participant made one eight-way choice for each of 20 questions.
StyleComposer received 200 of 600 selections (33.3\%).}
\label{fig:supp_user_study}
\end{figure}
\smallskip
\noindent\textbf{Transfer and leakage.}
Let $i,j\in\{c,t,s\}$ denote the source-reference role and measured attribute
axis, respectively, and let $D_j(I,R_i)$ be the corresponding primary
distance. For every cell, we first average its raw distances over the 250
three-reference cases, denoted by $\overline D_j$. As the common anchor without attribute conditioning, we use a content-only Flux output $I^p$ from the original prompt $p$, generated without image references or attribute descriptions. We compute
\begin{equation}
G_{ij}
=
\frac{\overline D_j(I^p,R_i)-\overline D_j(I,R_i)}
{\max\!\left(\overline D_j(I^p,R_i),10^{-6}\right)}.
\label{eq:supp_directional_gain}
\end{equation}
Thus, $G_{ii}$ measures target transfer, while $G_{ij}$ for $i\ne j$
measures movement toward source $i$ on a non-target axis. Negative movement counts as neither successful transfer nor leakage. We set
$t_i=[G_{ii}]_+$ and $\ell_i=\sum_{j\ne i}[G_{ij}]_+$, yielding
\begin{equation}
\mathrm{Selectivity}
=
\frac{1}{|\mathcal A|}
\sum_{i\in\mathcal A}
\frac{t_i}{t_i+\ell_i},
\label{eq:supp_selectivity}
\end{equation}
where a term is set to zero if its denominator vanishes. This construction prevents
a negative movement on one axis from canceling positive leakage on another.
\textit{Selectivity} measures only where the reference-directed change falls, so we report it together with \textit{Composition} and target transfer, which measure how large the change is. The full directional matrices of every method appear in
\cref{sec:supp_directional_matrices}.

\subsection{User Study Protocol}
\label{sec:supp_user_study}

Thirty participants answered 20 questions each, which gives 600 selections. The 20 three-reference cases cover both content prompts and every reference of each pool. Each question showed the references labeled Color, Texture, and Structure, the content prompt, and eight candidate outputs labeled A--H without method names.
We permuted the question order and the candidate positions once, and all participants saw the same anonymized form.
Participants selected the one image
that best incorporated all three designated references while following the
prompt. The instructions defined color as the overall palette, texture as
rendering characteristics such as medium, brushwork, and grain, and structure
as the coarse spatial arrangement of major regions.
After collection, a fixed key mapped the answer positions back to the method names, and \cref{fig:supp_user_study} gives the selections of every method.

\section{Representation Analysis}
\label{sec:supp_analysis}

\subsection{Attention Probes}
\label{sec:supp_attention_probes}

\begin{figure}[t]
\centering
\includegraphics[width=\linewidth]{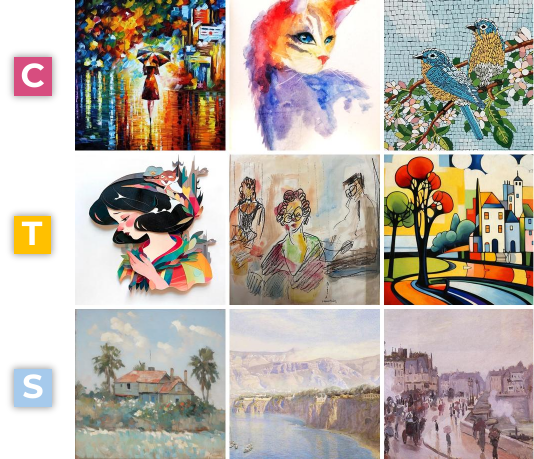}
\caption{\textbf{Reference images used in the attention probes.} Each row holds the three references selected mainly for color, texture, or spatial structure, and each is evaluated on all three attribute axes.}
\label{fig:supp_attention_refs}
\end{figure}

\smallskip
\noindent\textbf{Probe construction.}
As shown in \fref{fig:supp_attention_refs}, nine reference images are selected to cover diverse palettes, textures,
and structures, with three references chosen primarily for each
style attribute. Every output is evaluated against its reference on all three
attribute axes. Combining the nine references with five content prompts and
two seeds yields 90 outputs per condition. We use seeds 0 and 1 and the prompts
``A photo of a woman playing guitar'', ``A dog catching a
frisbee'', ``A lighthouse on a snowy shore'', ``A fruit basket on
a table'', and ``A photo of a city''.

To measure the raw attention routes, we turn off every restriction and schedule of the method, namely frequency filtering, statistics alignment,
the texture-attention cap, latent color alignment, block selection, and
temporal decay. Unless stated otherwise, each intervention acts on all
19 dual-stream and 38 single-stream attention blocks over all 28 sampling
steps. Only image tokens are recorded or routed; the text tokens remain those
of the generation prompt.

\begin{table}[t]
\centering
\footnotesize
\setlength{\tabcolsep}{5.3pt}
\caption{\textbf{Attention-probe results.}
Color, texture, and structure columns report signed movement toward the corresponding reference; CLIP-T reports prompt similarity. Each entry is the mean over 90 reference--prompt--seed combinations.}
\label{tab:supp_attention_matrix}
\begin{tabular*}{\linewidth}{@{\extracolsep{\fill}}lrrrr@{}}
\toprule
Condition & Color & Texture & Structure & CLIP-T \\
\midrule
Plain                         &  0.000 &  0.000 &  0.000 & 0.243 \\
\midrule
K only                        &  0.011 & -0.003 & -0.037 & 0.242 \\
V only                        &  0.528 &  0.395 &  0.138 & 0.121 \\
\midrule
K/V                           &  0.464 &  0.381 &  0.304 & 0.202 \\
K/V, K shuffled               &  0.322 &  0.025 & -0.085 & 0.211 \\
K/V, V neutralized            & -0.049 &  0.174 &  0.186 & 0.230 \\
\midrule
Q, $w{=}1$, steps 0--27       &  0.238 &  0.251 &  0.307 & 0.127 \\
Q, $w{=}0.5$, steps 0--27     &  0.029 &  0.049 &  0.012 & 0.179 \\
Q, $w{=}0.25$, steps 0--27    &  0.013 & -0.005 &  0.045 & 0.235 \\
Q, $w{=}1$, steps 0--5        &  0.312 &  0.083 & -0.030 & 0.229 \\
Q, $w{=}1$, steps 0--11       &  0.263 &  0.094 &  0.141 & 0.195 \\
Q, $w{=}1$, steps 0--17       &  0.244 &  0.132 &  0.108 & 0.162 \\
\bottomrule
\end{tabular*}
\end{table}

For the K/V probes, we encode the reference image $I^r$ into its VAE latent $z_0^r$.
At sampling step $i$, a Gaussian sample $\epsilon^r$ that is fixed within each generation brings the reference to the current noise level,
\begin{equation}
x_i^r=(1-\sigma_i)z_0^r+\sigma_i\epsilon^r.
\label{eq:supp_reference_noise}
\end{equation}

We forward $x_i^r$ at the same timestep as the generation using the neutral reference prompt ``a photo'', and record its image-token keys $K_i^r$ and values $V_i^r$. Raw K/V routing appends both tensors to the generation sequence. To change one component at a time, K only appends reference $K_i^r$ with a copy of generation $V_i^g$, whereas V only appends reference $V_i^r$ with a copy of generation $K_i^g$.

K shuffling applies one fixed random token permutation $\pi$ within each generation,
\begin{equation}
\widetilde K_{i,j}^r=K_{i,\pi(j)}^r,
\label{eq:supp_k_shuffle}
\end{equation}
while leaving $V_{i,j}^r$ in its original order. V neutralization instead replaces reference V with a copy whose statistics match those of generation V,
\begin{equation}
\widetilde V_i^r
=
\sqrt{\mathrm{var}(V_i^g)+\varepsilon}
\frac{V_i^r-\mu(V_i^r)}{\sigma(V_i^r)}
+\mu(V_i^g),
\label{eq:supp_v_neutralization}
\end{equation}
where the mean and standard deviation are computed over image tokens for
each attention head and channel, and
$\sigma(V_i^r)=\sqrt{\mathrm{var}(V_i^r)+\varepsilon}$ with
$\varepsilon=10^{-6}$.

For the Q probe, we record image-token queries from the reference image once
at scheduler index 20 using
$x_{\mathrm{anchor}}^r=(1-\sigma_{20})z_0^r$ and the neutral prompt
``a photo''. The recorded query is reused throughout sampling. Full Q
replacement applies $Q_i^g\leftarrow Q^r$ in all blocks and steps. The strength sweep instead applies $Q_i^g\leftarrow(1-w)Q_i^g+wQ^r$ for $w\in\{0.25,0.5\}$, and the duration
sweep applies full replacement only over steps 0--5, 0--11, or 0--17.

\begin{figure}[t]
\centering
\includegraphics[width=\linewidth]{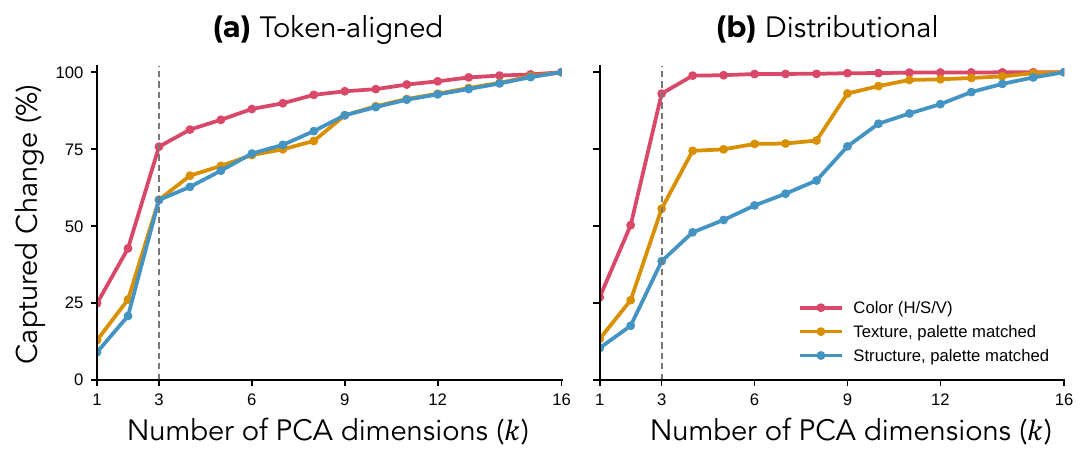}
\caption{\textbf{Cumulative latent-change capture along the PCA basis.} (a) Token-aligned change compares corresponding spatial tokens. (b) Distributional change compares empirical coordinate distributions and is the measure used in the main paper. Lines show means over edits, the dashed line marks the chromatic subspace, and texture and structure use palette-matched edits.}
\label{fig:supp_latent_capture}
\end{figure}

\smallskip
\noindent\textbf{Signed reference-relative movement.}
Let $I^0$ be the generation without injection, $I^q$ the output of probe $q$, and $R$ the reference. For color and structure distances $D_c$ and $D_s$, and texture similarity $S_t$, we compute
\begin{equation}
\Delta_c^q
=
\frac{D_c(I^0,R)-D_c(I^q,R)}
{\max\!\left(D_c(I^0,R),\varepsilon\right)},
\label{eq:supp_probe_color}
\end{equation}
\begin{equation}
\Delta_t^q
=
\frac{S_t(I^q,R)-S_t(I^0,R)}
{\max\!\left(1-S_t(I^0,R),\varepsilon\right)},
\label{eq:supp_probe_texture}
\end{equation}
\begin{equation}
\Delta_s^q
=
\frac{D_s(I^0,R)-D_s(I^q,R)}
{\max\!\left(D_s(I^0,R),\,0.5\,\widetilde D_s^0\right)},
\label{eq:supp_probe_structure}
\end{equation}
where $D_c$ is MS-SWD, $S_t$ is grayscale CSD similarity, and, for this
diagnostic analysis, $D_s$ is the mean-squared distance between per-image
standardized MiDaS-small~\cite{ranftl2022towards} depth maps. The main benchmark in our paper instead uses
Depth Anything V2~\cite{yang2024depth} as specified. The texture distance is
equivalently $1-S_t$. Furthermore, $\widetilde D_s^0$ is the median distance
between the no-injection outputs and their references, and
$\varepsilon=10^{-6}$ prevents division by zero. A positive value means movement toward the reference, zero means no change, and a negative value means movement away. CLIP-T is reported as a similarity, so higher is better.

\smallskip
\noindent\textbf{Results.}
Raw K/V routing transfers several reference properties at once. K alone has
little effect, whereas V strongly transfers color and texture but reduces
CLIP-T. Shuffling K suppresses texture while retaining color; neutralizing V
statistics suppresses color while retaining part of the texture. These
results show that texture depends more on token-wise K--V correspondence,
whereas color depends more on the global statistics of V. Neither can be
isolated by selecting K or V alone. These raw probes omit latent color
alignment and therefore measure the color that the attention pathway brings in before any correction. In the full method, we avoid additional statistical
transfer through V and use latent color alignment to correct the residual
palette transfer.

Under full replacement, Q produces its largest attribute movement on
structure, but reduces CLIP-T from 0.243 to 0.127. Lowering $w$ weakens
structure transfer, and shortening the interval does not consistently remove
color and texture changes. The sweeps therefore show that Q does not provide
structure-only control. 

\subsection{Latent Color Subspace}
\label{sec:supp_latent_subspace}

\begin{figure}[t]
\centering
\includegraphics[width=\linewidth]{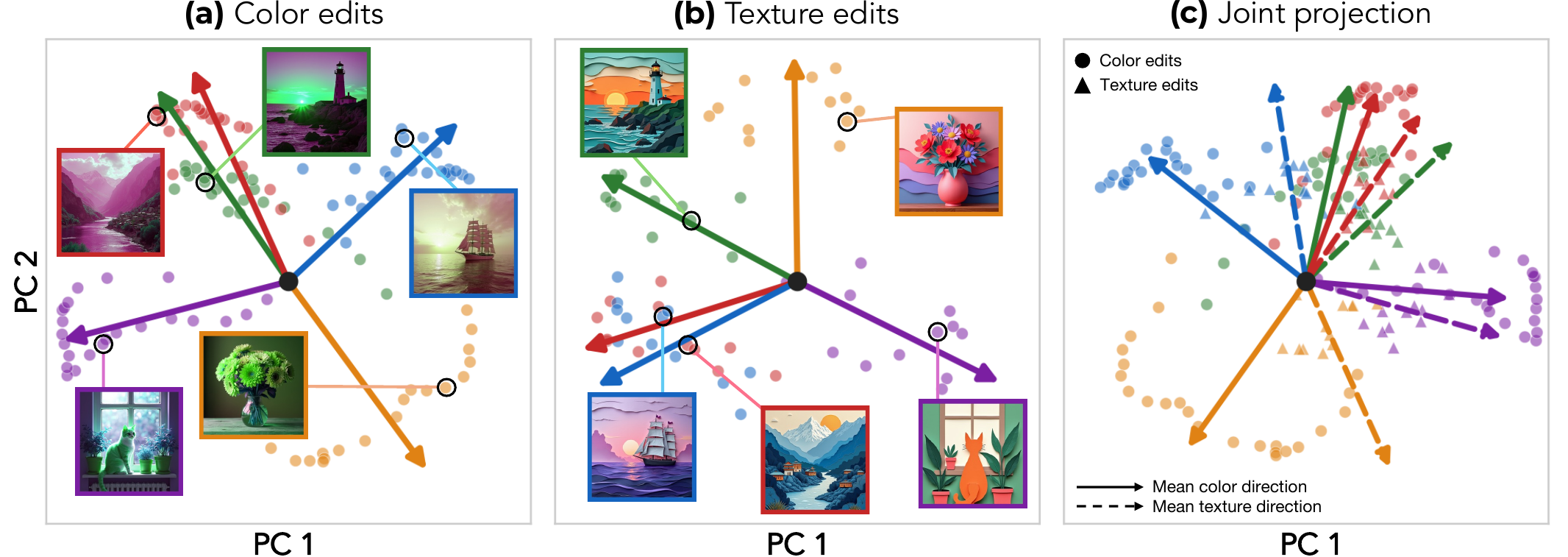}
\caption{\textbf{Edit directions in CLIP space.} Each marker is the unit displacement of one edit from its base image, the marker color identifies the base scene, and arrows give the mean direction of each scene. (a) and (b) project hue and texture edits, and (c) projects both together. Directions cluster by base scene, and continuous hue edits trace curved trajectories that differ between scenes.}
\label{fig:embedding}
\end{figure}
\smallskip
\noindent\textbf{Basis and edit construction.}
Using a fixed random seed of 0, we sample 512 RGB colors uniformly, render
each as a $512\times512$ solid-color image, and encode it with the Flux VAE.
We average each latent over spatial positions and apply PCA to the resulting
512 vectors in 16-D embedding. The first three components explain 48.9\%,
32.7\%, and 15.6\% of the variance, respectively, or 97.3\% cumulatively.
This defines the 3-D chromatic basis used below. We then
measure how much of each color, texture, and structure edit falls within it.

The edits start from the five base scenes of \cref{sec:supp_clip}. Color edits rotate the hue by $45^\circ$, $135^\circ$, $225^\circ$, or $315^\circ$, scale the saturation by 0.50, 0.75, 1.25, or 1.50, or scale the value by 0.60, 0.80, 1.20, or 1.40.
This gives 60 color pairs, with equal counts for the three variables. Texture edits use the 12 style variants from
the same five base scenes, giving 60 pairs. Structure edits apply a
horizontal flip, horizontal and vertical cyclic shifts by one quarter of the
image size, and a $1.3\times$ center zoom, giving 20 pairs.

To keep the palette of the non-color edits fixed, we match each RGB marginal
of every texture and structure edit to that of its base image using
rank-preserving histogram specification. The mean RGB-histogram
total-variation distance is reduced from 0.306 to 0 for texture and from
0.039 to 0 for structure.  We use these palette-matched variants for the texture and structure curves below.

\smallskip
\noindent\textbf{Captured change.}
Let $X,X'\in\mathbf{R}^{N\times16}$ be the base and edited latent tokens, and let $U=[u_1,\ldots,u_{16}]$ be the PCA basis ordered by explained variance.
Since the color route matches distributions over positions, the change along each basis dimension is the squared one-dimensional Wasserstein distance between the sorted coordinates,
\begin{equation}
E_\ell
=
\frac{1}{N}
\left\|
\mathrm{sort}(X'u_\ell)-\mathrm{sort}(Xu_\ell)
\right\|_2^2,
\label{eq:supp_distribution_energy}
\end{equation}
and the fraction captured by the first $k$ dimensions is
\begin{equation}
r_k
=
\frac{\sum_{\ell=1}^{k}E_\ell}
{\sum_{\ell=1}^{16}E_\ell}.
\label{eq:supp_distribution_capture}
\end{equation}
Because $r_k$ is normalized within each edit, it measures how concentrated the change is across dimensions, and a change spread equally over all 16 dimensions would put 18.75\% in the first three.
The token-aligned variant in \cref{fig:supp_latent_capture}a uses the coordinates of corresponding tokens without sorting.
Curves average over the edits of each family.

\begin{table}[t]
\centering
\footnotesize
\setlength{\tabcolsep}{2.5pt}
\caption{\textbf{Variable base-prompt fields in the CLIP-space analysis.} All scenes share the base style ``high quality photograph''.}
\label{tab:supp_clip_prompts}
\begin{tabularx}{\linewidth}{@{}l>{\raggedright\arraybackslash}X@{}}
\toprule
Scene & Prompt fields \\
\midrule
Lighthouse &
\emph{Scene:} a lighthouse on a rocky shore at sunset\newline
\emph{Palette:} teal and orange palette \\[3pt]
Cat &
\emph{Scene:} a cat sitting by a window with potted plants\newline
\emph{Palette:} green and pink palette \\[3pt]
Village &
\emph{Scene:} a mountain village beside a river\newline
\emph{Palette:} blue and gold palette \\[3pt]
Flowers &
\emph{Scene:} a bouquet of flowers in a vase on a wooden table\newline
\emph{Palette:} red and violet palette \\[3pt]
Ship &
\emph{Scene:} a sailing ship on the open sea\newline
\emph{Palette:} purple and silver palette \\
\bottomrule
\end{tabularx}
\end{table}
\begin{table}[t]
\centering
\footnotesize
\setlength{\tabcolsep}{2.8pt}
\caption{\textbf{Robustness of content-dependent edit directions.} The pairwise column gives the range of the ten cosines between scene directions, whose mean is $\kappa^a$, and the leave-one-content-out (LOO) range recomputes $\kappa^a$ after excluding each scene in turn. The last column gives the variance explained by the 2-D PCA view of \fref{fig:embedding}, and all cosines use the original 512-D embeddings.}
\label{tab:supp_clip_robustness}
\begin{tabular*}{\linewidth}{@{\extracolsep{\fill}}lcccc@{}}
\toprule
Edit
& $\kappa^a$
& Pairwise cos. range
& LOO $\kappa^a$ range
& PCA variance \\
\midrule
Color            & 0.24 & 0.03--0.46 & 0.18--0.31 & 38\% \\
Texture & 0.37 & 0.20--0.58 & 0.29--0.44 & 31\% \\
\bottomrule
\end{tabular*}
\end{table}
\smallskip
\noindent\textbf{Results.}
At $k=3$, color edits reach 93.0\% distributional capture, whereas palette-matched texture and structure edits reach 55.6\% and 38.6\% (\cref{fig:supp_latent_capture}b).
The color curve flattens after the third dimension, and the texture and structure curves keep rising well beyond it.
The token-aligned curves give the same ordering at $k=3$.
Color change therefore concentrates in the three chromatic dimensions, which are the coordinates that the color route moves.

\subsection{Image Embedding}
\label{sec:supp_clip}

Adapter-based stylization methods~\citep{ye2023ip} compress a reference image into the CLIP embedding, and SADis~\citep{qin2025free} separates color from texture by arithmetic in that same space, using image-specific differences and an additional color-correction stage. For such control to be meaningful, an attribute change must correspond to a direction that is reusable across content; otherwise the same strength value does not have the same meaning across images. We analyze this in the 512-D pooled embedding of CLIP ViT-B/32~\citep{radford2021learning}.

From five fixed-seed scenes we apply 23 post-hoc hue rotations per scene, which preserve content and layout exactly, and separately regenerate 12 texture variants by changing only the style phrase while fixing the scene content, palette phrase, and seed.
Let $e(\cdot)$ denote the L2-normalized CLIP image embedding; for each edit we take its unit displacement $d = (e(I')-e(I))/\|e(I')-e(I)\|_2$ and average the displacements of one scene into that scene's mean direction $m$.
The two edit families are treated separately, and we average the ten pairwise cosine similarities among the mean directions of the five scenes.

A reusable, content-independent direction would produce a value near one, whereas unrelated directions have an expected cosine near zero. The measured similarities are 0.24 for color and 0.37 for texture edits, placing both substantially closer to the unrelated-direction baseline than to perfect alignment. In \fref{fig:embedding}, the displacements group by base scene rather than along a shared per-attribute axis. CLIP embedding summarizes scene content and appearance jointly, so the same nominal edit interacts with the underlying scene and produces a different displacement. We therefore do not route any attribute through this representation.

\smallskip
\noindent\textbf{Data construction.}
We first generate five base scenes with Flux.1-dev~\cite{flux2024} at $512\times512$ resolution, using 28 sampling steps, guidance scale 3.5, and seed 42. Each base prompt combines the scene description, ``high quality photograph'', and a fixed palette phrase (\tref{tab:supp_clip_prompts}).

For color, we rotate HSV hue of each base image in $15^\circ$ increments from $15^\circ$ to $345^\circ$, yielding 23 outputs per scene. For texture, we keep the scene description, palette phrase, seed, and sampling settings fixed and replace only the style phrase. The 12 analyzed styles are impasto, mosaic, felt, paper cutout, stained glass, airbrush, origami, embroidery, claymation, low-poly rendering, ukiyo-e, and graffiti.

\smallskip
\noindent\textbf{Consistency and robustness.}
We encode all images using the 512-dimensional pooled image embedding of
CLIP ViT-B/32~\cite{radford2021learning}. Using the unit displacement
$d_{i,j}^{a}$ defined above, we normalize the mean edit direction
for content $i$ as
\begin{equation}
m_i^{a}
=
\frac{\sum_j d_{i,j}^{a}}
{\left\|\sum_j d_{i,j}^{a}\right\|_2},
\label{eq:supp_content_direction}
\end{equation}
and compute cross-content consistency as
\begin{equation}
\kappa^{a}
=
\frac{1}{\binom{5}{2}}
\sum_{i<k}
\left(m_i^{a}\right)^\top m_k^{a}.
\label{eq:supp_direction_consistency}
\end{equation}

As \fref{fig:embedding} and \tref{tab:supp_clip_robustness}, the leave-one-content-out (LOO) ranges remain far below perfect alignment, showing that no single
scene determines the conclusion.

\section{Route Implementation}
\label{sec:supp_method}

\subsection{Sampling Algorithm}
\label{sec:supp_sampling}
\begin{algorithm}[t]
\caption{StyleComposer sampling}
\label{alg:supp_stylecomposer}
\begin{algorithmic}[1]
\REQUIRE Prompt $p$; references $I_c,I_t,I_s$; strengths
$\alpha_c,s_{\mathrm{lf}},T_s$; noise schedule
$\{\sigma_i\}_{i=0}^{N}$
\STATE Encode active references:
$z_0^c\leftarrow\mathrm{VAEEnc}(I_c)$,
$z_0^t\leftarrow\mathrm{VAEEnc}(I_t)$,
$z_0^s\leftarrow\mathrm{VAEEnc}(I_s)$
\STATE Sample initial latent $x_0$ and texture-reference noise $\epsilon^t$
\STATE Record structure-query anchor
$Q^s\leftarrow\mathrm{RecordQ}(z_0^s)$
\FOR{$i=0,\ldots,N-1$}
    \STATE Initialize the frozen transformer with generation tokens from
    $(x_i,p)$
    \IF{$i$ is in the structure interval}
        \STATE Blend image queries with $Q^s$ by cosine weight $w_i$
    \ENDIF
    \IF{$i$ is in the texture interval}
        \STATE $x_i^t\leftarrow
        (1-\sigma_i)z_0^t+\sigma_i\epsilon^t$
        \STATE Extract $(K_i^t,V_i^t)$ from $x_i^t$ at the current timestep
        \STATE Align generation-key statistics to $K_i^t$
        \STATE Retain low-frequency $K_i^t$ and append reference image-token
        K/V under the attention cap
    \ENDIF
    \STATE Predict routed velocity
    $v_i\leftarrow v_\theta(x_i,\sigma_i,p)$
    \STATE Form clean estimate
    $\hat{x}_0\leftarrow x_i-\sigma_i v_i$
    \IF{$i$ is in the color interval}
        \STATE Predict reference-free $\hat{x}_0^{\mathrm{base}}$ from $(x_i,\sigma_i,p)$
        \STATE Project $\hat{x}_0$, $\hat{x}_0^{\mathrm{base}}$, and $z_0^c$
        onto the chromatic basis
        \STATE Transport the chromatic distribution to that of $z_0^c$
        \STATE Interpolate transported and base coordinates by $\alpha_c$
        \STATE Lift only the chromatic displacement to obtain
        $\hat{x}_0^{\,\prime}$
        \STATE $v_i\leftarrow
        (x_i-\hat{x}_0^{\,\prime})/\sigma_i$
    \ENDIF
    \STATE $x_{i+1}\leftarrow
    x_i+(\sigma_{i+1}-\sigma_i)v_i$
\ENDFOR
\RETURN $\mathrm{VAEDec}(x_N)$
\end{algorithmic}
\end{algorithm}

Algorithm~\ref{alg:supp_stylecomposer} summarizes the complete sampling
procedure. The pretrained Flux transformer, VAE, text encoders, and scheduler
remain frozen. We first encode the three references into VAE latents. The
structure reference is forwarded once at a fixed noise level to record its
image-token queries. Texture features are instead recomputed at the current
noise level during every active texture step. The color reference needs no transformer forward pass, because its latent is projected directly onto the chromatic basis.

In each sampling step, the structure and texture routes act inside the transformer pass that predicts the velocity. We convert this velocity into a clean latent estimate and align only its chromatic coordinates during the active color steps. Color alignment also uses a reference-free clean
prediction from the same noisy latent. This prediction gives the $\alpha_c=0$ endpoint, so the palette that an attention route brings in does not become the baseline of the color slider. The aligned clean estimate is
converted back to velocity before applying the original scheduler update.
Thus, all three routes alter intermediate features or the clean prediction,
while the pretrained sampler itself is unchanged.

\subsection{Color Route}
\label{sec:supp_color_route}

Let $B\in\mathbf{R}^{16\times3}$ and $\mu\in\mathbf{R}^{16}$ denote the
orthonormal chromatic basis and latent mean, respectively. Both are
precomputed once and shared across all prompts and references. For a clean
latent estimate $\hat{x}_0$, we flatten its spatial locations and obtain
chromatic coordinates
\begin{equation}
C=(\hat{X}_0-\mu)B,\qquad
C^c=(Z_0^c-\mu)B,
\label{eq:supp_color_coordinates}
\end{equation}
where $\hat{X}_0,Z_0^c\in\mathbf{R}^{N\times16}$. The reference coordinate set $C^c$ is computed once, and only the generated coordinates change over the sampling steps.

\smallskip
\noindent\textbf{Sliced distribution transport.}
We implement the transport $\mathcal{T}(C,C^c)$ with four iterative random
rotations. At iteration $r$, an orthonormal matrix
$R_r\in\mathbf{R}^{3\times3}$ rotates both coordinate sets. For each rotated
axis, the generated coordinates are sorted and assigned the corresponding
quantiles of the reference coordinates. The three matched axes are then rotated back, which gives the coordinates for the next iteration,
\begin{equation}
C^{(r+1)}
=
\mathcal{H}\!\left(C^{(r)}R_r,\ C^cR_r\right)R_r^\top,
\label{eq:supp_sot_iteration}
\end{equation}
where $\mathcal{H}$ matches empirical quantiles independently along the three columns and $C^{(0)}=C$. The rotations let the matching account for the correlations among the three coordinates. We finally expand the transported distribution around the reference mean by a fixed factor $\rho=1.5$, so that
\begin{equation}
\mathcal{T}(C,C^c)
=
\bar{C}^c+\rho\left(C^{(4)}-\bar{C}^c\right).
\label{eq:supp_color_spread}
\end{equation}
Here, $\bar C^c$ denotes the reference-coordinate mean, broadcast over
spatial positions. We sample four rotations independently for each
application; their pseudo-random stream is independent of the diffusion seed.

\smallskip
\noindent\textbf{Base interpolation and lifting.}
Let $C^{\mathrm{base}}$ be the chromatic coordinates of a clean prediction from the same noisy latent with all reference routes disabled, and let the color target be
\begin{equation}
\bar{C}
=
(1-\alpha_c)C^{\mathrm{base}}
+\alpha_c\mathcal{T}(C,C^c).
\label{eq:supp_color_interpolation}
\end{equation}
Thus, $\alpha_c=0$ recovers the palette of the reference-free prediction, and the color route also removes the palette brought in by reference V. We lift only the coordinate displacement back to the 16 latent channels,
\begin{equation}
\hat{X}_0^{\,\prime}
=
\hat{X}_0+(\bar{C}-C)B^\top.
\label{eq:supp_color_lift}
\end{equation}
Because $B$ has orthonormal columns, this update lies entirely in the chromatic subspace and leaves the projection of $\hat{X}_0$ onto the 13-dimensional orthogonal complement unchanged.

\begin{table}[t]
\centering
\footnotesize
\setlength{\tabcolsep}{3pt}
\caption{\textbf{StyleComposer configuration.}
Every experiment uses these settings unless its section states otherwise.}
\label{tab:supp_hyperparameters}
\begin{tabularx}{\linewidth}{@{}l>{\raggedright\arraybackslash}X@{}}
\toprule
Component & Final setting \\
\midrule
Backbone & Flux.1-dev, bfloat16 \\
Resolution & $512\times512$ \\
Sampling & 28 steps, guidance 3.5 \\
Scheduler & FlowMatch Euler, sequence-length-dependent shift \\
Benchmark seed & 42 for every case \\
Text length & Maximum 512 tokens \\
Generation prompt & Content prompt plus the texture-medium descriptor when
texture is active \\
Color route & Steps 16--27; $\alpha_c=1$ \\
Color transport & 4 rotations; 12 one-dimensional matches;
$\rho=1.5$ \\
Texture route & Steps 3--27; all 19 dual- and 38 single-stream blocks \\
Texture frequency & $s_{\mathrm{hf}}=0$, $s_{\mathrm{lf}}=2$,
$\beta=2$ \\
Texture statistics & AdaIN on generation K only \\
Texture cap & 0.25 for steps 3--11; 0.50 for 12--19; none for 20--27 \\
Structure route & Steps 0--11; last 25 single-stream blocks;
$T_s=12$ \\
Structure anchor & Scheduler index 20; initial weight 1; cosine decay \\
Reference prompt & \texttt{a photo} \\
\bottomrule
\end{tabularx}
\end{table}

In the final configuration, color routing is applied at steps 16--27 with
$\alpha_c=1$. Beyond projection, sorting, and lifting, its only model
evaluation is one additional reference-free transformer forward at each
active step. No gradient, learned color module, or inversion is used.
Matching a generated color distribution to a reference during sampling has precedents that transform the covariance of latent codes~\cite{agarwal2024training} or guide sampling toward a target color histogram by optimal transport~\cite{liu2026histogram}, and closed-form edits toward a target color already use the latent color subspace~\cite{pach2026latent}.
The color route moves only three latent coordinates per position and works without decoding the latent into an image.

\subsection{Texture Route}
\label{sec:supp_texture_route}

At every active texture step, we match the encoded texture reference to the
current scheduler level using the same construction as
\eref{eq:supp_reference_noise}. A neutral reference prompt,
``a photo'', is used when extracting its image-token features. Texture
features are routed through all 19 dual-stream and 38 single-stream attention
blocks over steps 3--27.

\smallskip
\noindent\textbf{Low-frequency keys and K-only alignment.}
Let a RoPE axis contain $n$ frequency pairs, indexed by $j=0,\dots,n-1$ from high to low frequency. Except for the first positional axis, we scale frequency pair $j$ by
\begin{equation}
m_j
=
s_{\mathrm{hf}}
+
(s_{\mathrm{lf}}-s_{\mathrm{hf}})
\left(\frac{j}{n-1}\right)^\beta.
\label{eq:supp_texture_frequency}
\end{equation}

The first axis is scaled uniformly by $s_{\mathrm{lf}}$. The final
configuration uses $s_{\mathrm{hf}}=0$, $s_{\mathrm{lf}}=2$, and
$\beta=2$. Consequently, fine positional frequencies are suppressed while
coarser correspondence is retained and progressively emphasized.

\begin{table}[t]
\centering
\footnotesize
\setlength{\tabcolsep}{3pt}
\caption{\textbf{Inference time on a single NVIDIA H100 80GB GPU.}
Times exclude model initialization. The B-LoRA inference time excludes per-reference
optimization.}
\label{tab:supp_runtime}
\begin{tabular*}{\linewidth}{@{\extracolsep{\fill}}lrlr@{}}
\toprule
Method & sec/image & Method & sec/image \\
\midrule
Flux & 5.81 & SDXL & 7.48 \\
InstantStyle & 5.81 & B-LoRA & 7.57 \\
SADis & 7.99 & StyleAligned & 22.56 \\
IP-Adapter Flux & 16.51 & StyleComposer (ours) & 10.33 \\
\bottomrule
\end{tabular*}
\end{table}
Before RoPE, we align the image-token statistics of generation keys to the texture-reference keys independently for every attention head and channel with the update
\begin{equation}
K_g^{\mathrm{img}}
\leftarrow
\sqrt{\mathrm{var}(K^t)+\varepsilon}
\frac{K_g^{\mathrm{img}}-\mu(K_g^{\mathrm{img}})}
{\sqrt{\mathrm{var}(K_g^{\mathrm{img}})+\varepsilon}}
+\mu(K^t),
\label{eq:supp_texture_adain}
\end{equation}
where statistics are computed over image tokens and $\varepsilon=10^{-6}$. We do not align Q or V.
After RoPE and the frequency scale, the reference keys $\widetilde K^t$ and the unmodified reference values $V^t$ are appended to the image tokens of the generation. K-only alignment makes the two key sets comparable
without introducing an additional V-side statistics transfer. The color route overwrites the palette that reference V still brings in during its active steps.

\smallskip
\noindent\textbf{Reference-attention cap.}
Early in sampling, unrestricted reference tokens can dominate the
prompt-conditioned generation. We therefore bound, for each generated query, the total attention mass on the appended texture tokens. Let
\begin{equation}
\begin{aligned}
L_i^t
&=
Q_i(\widetilde K_i^t)^\top/\sqrt d,\\
L_i^g
&=
Q_i(K_i^g)^\top/\sqrt d,\\
a_i
&=
\mathrm{sigmoid}\!\left(
\mathrm{LSE}(L_i^t)-\mathrm{LSE}(L_i^g)
\right).
\end{aligned}
\label{eq:supp_texture_mass}
\end{equation}
be the reference mass implied by the two key sets, where $\mathrm{LSE}$ is
taken over the key-token dimension. We compute attention
outputs $O_i^g$ and $O_i^t$ over generation and reference tokens separately
and mix them as
\begin{equation}
O_i=(1-\widetilde a_i)O_i^g+\widetilde a_iO_i^t,
\qquad
\widetilde a_i=\min(a_i,\tau_i).
\label{eq:supp_texture_cap}
\end{equation}
Within the active texture interval, the cap is $\tau_i=0.25$ for
$3\le i<12$, $\tau_i=0.50$ for $12\le i<20$, and is removed for
$i\ge20$. This schedule preserves prompt-driven content formation early and
allows stronger texture accumulation later. The texture slider changes
$s_{\mathrm{lf}}$ while leaving the route interval and cap schedule fixed.

\subsection{Structure Route}
\label{sec:supp_structure_route}

\begin{table}[t]
\centering
\footnotesize
\setlength{\tabcolsep}{3pt}
\caption{\textbf{Route settings.} Each row changes one setting from the benchmark setting on 40 cases. The benchmark setting records the structure anchor once at scheduler index 20 without noise, blends it into the last 25 single-stream blocks for $T_s=12$ steps, and extracts the texture keys and values at every step. The color, texture, and structure distances are MS-SWD, grayscale CSD, and one minus the depth correlation. Re-extracting the anchor at every step returns the depth distance to the level without a structure reference.}
\label{tab:supp_route_settings}
\begin{tabular*}{\linewidth}{@{\extracolsep{\fill}}lcccc@{}}
\toprule
Setting & MS-SWD$\downarrow$ & gCSD$\downarrow$ & Depth$\downarrow$ & 1$-$CLIP-T$\downarrow$ \\
\midrule
Benchmark setting & 5.04 & 0.411 & 0.161 & 0.771 \\
Without structure reference & 5.12 & 0.461 & 0.310 & 0.754 \\
\midrule
Anchor at index 14 & 4.99 & 0.408 & 0.197 & 0.770 \\
Anchor at index 26 & 4.98 & 0.399 & 0.182 & 0.774 \\
Anchor at every step & 4.91 & 0.395 & 0.313 & 0.781 \\
Gaussian noise in anchor & 5.03 & 0.409 & 0.142 & 0.770 \\
First 25 blocks & 4.98 & 0.395 & 0.149 & 0.782 \\
$T_s=6$ & 5.02 & 0.444 & 0.184 & 0.759 \\
$T_s=18$ & 5.04 & 0.411 & 0.234 & 0.792 \\
$T_s=24$ & 5.19 & 0.441 & 0.240 & 0.818 \\
\midrule
Texture K/V at index 20 & 5.02 & 0.412 & 0.186 & 0.785 \\
\bottomrule
\end{tabular*}
\end{table}
The structure route records its query anchor once. We encode the structure reference as
$z_0^s$ and form
\begin{equation}
x_{\mathrm{anchor}}^s=(1-\sigma_{20})z_0^s.
\label{eq:supp_structure_anchor}
\end{equation}
Forwarding this latent at scheduler index 20 with the neutral prompt
``a photo'' records its image-token queries $Q^{s,(\ell)}$. No Gaussian
noise is added to the anchor. We retain queries from the last 25 of the 38
single-stream blocks (indices 13--37) and reuse them throughout the active structure interval.

At step $i$, the generation queries of each selected block $\ell$ are mixed with the anchor as
\begin{equation}
\begin{aligned}
Q_i^{g,(\ell)}
&\leftarrow
(1-w_i)Q_i^{g,(\ell)}
+w_iQ^{s,(\ell)},\\
w_i
&=
\frac{1}{2}
\left[
1+\cos\left(\frac{\pi i}{T_s}\right)
\right].
\end{aligned}
\label{eq:supp_structure_decay}
\end{equation}
for $0\le i<T_s$; we set $w_i=0$ afterward. Only generation image-token
queries in the selected blocks are blended; text-token queries and all K/V
features remain unchanged by the structure route. The final benchmark uses
$T_s=12$ and an initial weight $w_0=1$. The anchor therefore has its largest
effect during early layout formation and is released before late appearance
and detail synthesis. Changing $T_s$ provides the structure-strength control;
$T_s=0$ disables this route. Since the anchor is recorded once, structure
routing adds no per-step reference forward pass.

\subsection{Settings and Runtime}
\label{sec:supp_hyperparameters}

\tref{tab:supp_hyperparameters} lists the fixed configuration used for
all single-, paired-, and three-reference benchmark cases. No setting is
selected per prompt or reference combination.

We measure the wall-clock time on one NVIDIA H100 80GB GPU after model initialization, and each value in \tref{tab:supp_runtime} averages 20 three-reference benchmark generations. StyleComposer takes 10.33 seconds per image and prompt-only Flux takes 5.81 seconds. The extra time comes from the forward passes of the texture reference and the reference-free forward passes of the color route, while the structure anchor is computed only once.

\subsection{Route Settings}
\label{sec:supp_route_settings}

\begin{table*}[!t]
\centering
\footnotesize
\setlength{\tabcolsep}{10pt}
\caption{\textbf{\textit{Composition} across all benchmark tiers.}
C, T, and S denote color, texture, and structure references.
A dash indicates that none of the requested references can enter the
method's image-conditioning mechanism. All scores use the normalization
fixed from the eight main methods; higher is better. Best and second-best
results are bold and underlined.}
\label{tab:supp_tier_composition}
\begin{tabular*}{\textwidth}{@{\extracolsep{\fill}}lccccccc@{}}
\toprule
Method
& C & T & S
& C+T & C+S & T+S
& C+T+S \\
\midrule
Flux & 0.648 & 0.391 & 0.716 & 0.354 & \underline{0.662} & 0.392 & 0.347 \\
SDXL & 0.650 & 0.473 & 0.522 & 0.418 & 0.510 & 0.352 & 0.407 \\
\midrule
InstantStyle & -- & 0.542 & -- & 0.263 & -- & 0.454 & 0.243 \\
StyleAligned & -- & 0.490 & -- & 0.362 & -- & 0.386 & 0.274 \\
IP-Adapter & -- & 0.354 & \underline{0.761} & 0.266 & 0.586 & 0.407 & 0.303 \\
SADis & \underline{0.675} & \underline{0.597} & -- & \underline{0.550} & 0.515 & \underline{0.504} & \underline{0.432} \\
B-LoRA & -- & 0.312 & -- & 0.306 & -- & 0.231 & 0.301 \\
\midrule
StyleComposer & \textbf{0.859} & \textbf{0.623} & \textbf{0.770}
& \textbf{0.654} & \textbf{0.845} & \textbf{0.586}
& \textbf{0.621} \\
\bottomrule
\end{tabular*}
\end{table*}
\begin{table*}[!t]
\centering
\footnotesize
\setlength{\tabcolsep}{8pt}
\caption{\textbf{Effect of the texture-medium descriptor} on the 250 three-reference cases. The image references, routes, content prompts, and seeds are otherwise identical, and CLIP-T always uses the content prompt without the descriptor.}
\label{tab:supp_texture_prompt}
\begin{tabular*}{\textwidth}{@{\extracolsep{\fill}}lcccccc@{}}
\toprule
Prompt configuration
& \textit{Composition}$\uparrow$ & \textit{Selectivity}$\uparrow$
& MS-SWD$\downarrow$ & gCSD$\downarrow$
& Depth$\downarrow$ & 1$-$CLIP-T$\downarrow$ \\
\midrule
Content prompt only
& 0.562 & 0.701 & \textbf{5.09} & 0.548 & \textbf{0.073} & \textbf{0.779} \\
+ texture-medium descriptor
& \textbf{0.621} & \textbf{0.745} & 5.32 & \textbf{0.439} & 0.088 & 0.785 \\
\bottomrule
\end{tabular*}
\end{table*}
\begin{table*}[!t]
\centering
\footnotesize
\setlength{\tabcolsep}{8pt}
\caption{\textbf{Naive modular composition on 250 three-reference cases.}
Indented rows apply fixed post-hoc color transfer to the IP-Adapter + Depth outputs, and SADis + Canny adds image-based structure control to SADis. Best and second-best results are bold and underlined.}
\label{tab:supp_naive_composition}
\begin{tabular*}{\textwidth}{@{\extracolsep{\fill}}lcccccc@{}}
\toprule
Method & \textit{Composition}$\uparrow$ & \textit{Selectivity}$\uparrow$
& MS-SWD$\downarrow$ & gCSD$\downarrow$
& Depth$\downarrow$ & 1$-$CLIP-T$\downarrow$ \\
\midrule
IP-Adapter + Depth & 0.303 & 0.370 & 10.98 & 0.487 & 0.256 & \underline{0.813} \\
\quad + Reinhard & \underline{0.330} & 0.434 & \underline{3.55} & 0.503 & 0.257 & 0.823 \\
\quad + Chroma Reinhard & 0.323 & 0.423 & 7.60 & 0.487 & 0.249 & 0.822 \\
\quad + Histogram Matching & 0.315 & 0.432 & \textbf{3.15} & 0.528 & 0.249 & 0.824 \\
SADis + Canny & 0.267 & \underline{0.600} & 9.08 & \underline{0.468} & \textbf{0.065} & 0.828 \\
\midrule
StyleComposer & \textbf{0.621} & \textbf{0.745} & 5.32 & \textbf{0.439} & \underline{0.088} & \textbf{0.785} \\
\bottomrule
\end{tabular*}
\end{table*}
We change one setting of the structure and texture routes at a time on the 40 cases of \cref{sec:supp_strength}, keeping all strengths at the benchmark setting (\cref{tab:supp_route_settings}).
Recording the structure anchor once matters most, because re-extracting the structure queries at every step raises the depth distance from 0.161 to 0.313, the level of the output without a structure reference.
Recording the anchor at scheduler index 14 or 26 also raises the depth distance, and adding Gaussian noise to the anchor gives 0.142, close to the noise-free anchor.
Blending the anchor into the first 25 single-stream blocks transfers the layout as well, whereas the last 25 blocks keep the output closer to the prompt.
Among structure windows $T_s$ of 6, 12, 18, and 24 steps with $w_0=1$, $T_s=12$ gives the lowest depth distance, the 6-step window gives the highest prompt similarity, and longer windows lower prompt similarity without bringing the layout closer.
For the texture route, recording the reference keys and values only once, at scheduler index 20, leaves the texture distance unchanged and raises both the depth and the prompt distance, and the route therefore extracts them at every step.

\begin{figure*}[t]
\centering
\includegraphics[width=\textwidth]{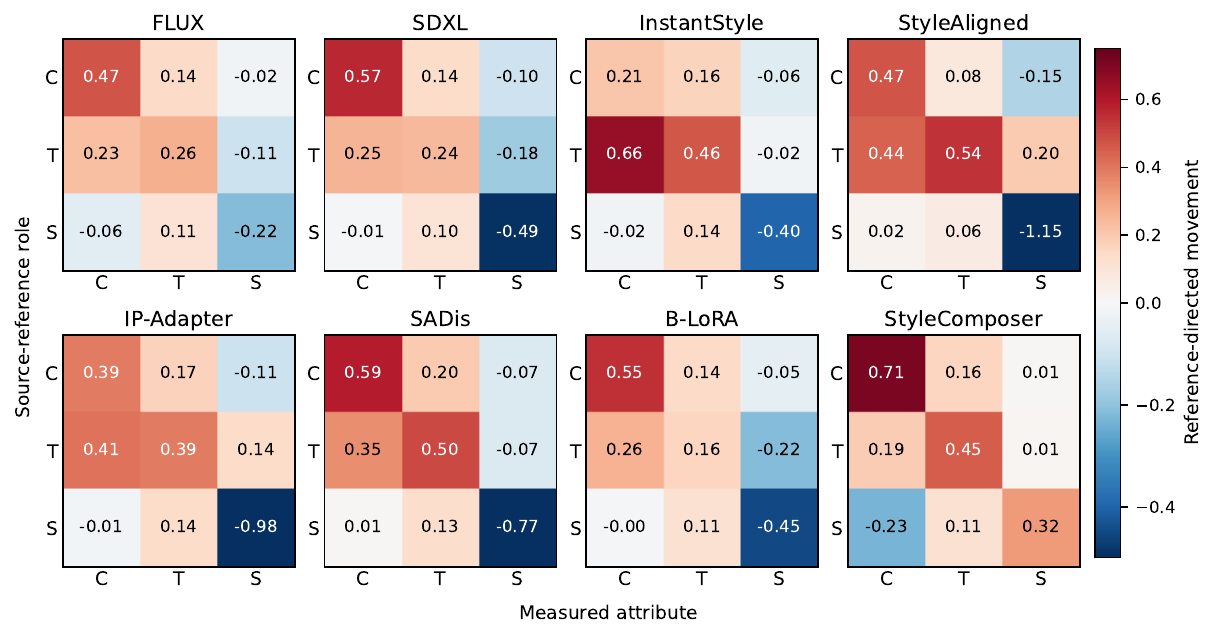}
\caption{\textbf{Complete reference-directed movement matrices.} Rows are source references and columns are measured attributes, with C, T, and S for color, texture, and structure. Diagonal cells are target transfer, and positive off-diagonal cells are cross-attribute leakage. Negative values denote movement away from the corresponding reference, and \textit{Selectivity} counts them as neither transfer nor leakage.}
\label{fig:supp_directional_matrices}
\end{figure*}

\section{Additional Comparisons}
\label{sec:supp_comparison}

\subsection{Results by Reference Combination}
\label{sec:supp_tier_results}

\tref{tab:supp_tier_composition} separates the aggregate benchmark according to the number and identity of requested references. StyleComposer achieves the
highest \textit{Composition} in every supported tier, from single-attribute transfer
to three-reference composition. The same fixed routes therefore keep the advantage for every subset of requested attributes.

\subsection{Without the Medium Descriptor}
\label{sec:supp_texture_prompt}

The final benchmark prompt appends a short medium descriptor associated with
the texture reference. To isolate its effect, we compare the final model
with an otherwise identical configuration using only the original content
prompt.

Without the descriptor, StyleComposer reaches a \textit{Composition} of 0.562 and a \textit{Selectivity} of 0.701, above every compared method in \cref{tab:main}.
The descriptor raises the texture fidelity further, and prompt alignment is always evaluated against the original content prompt.

\subsection{Combining Single-Attribute Controls}
\label{sec:supp_naive_composition}

We test whether existing controls can be composed without the routes of StyleComposer. The first alternative starts from the
IP-Adapter with depth ControlNet baseline, which receives the texture and
structure images, and applies a deterministic color-transfer operator to its
final output using the color reference. We test mean-and-standard-deviation
matching over all CIE Lab channels (Reinhard), the same matching over only
the chroma channels, and independent Lab-channel histogram matching. The second alternative augments SADis, which receives the color and texture images,
with the official SDXL Canny ControlNet driven by the structure reference. We
use the official Canny thresholds 50/100 and conditioning scale 0.4. Each configuration is fixed for all 250 three-reference cases, and no image receives its own selection or tuning.

The modular baselines can be strong on an individual axis. Histogram
matching attains the lowest color distance, and SADis + Canny attains the
lowest depth distance. These gains do not reach balanced composition, and their \textit{Composition} stays at or below 0.330.
StyleComposer has the highest \textit{Composition} and \textit{Selectivity} and the best texture and prompt alignment in this comparison.
Stacking strong single-axis controls therefore does not satisfy all three references and the content prompt at once.

\subsection{Directional Transfer and Leakage}
\label{sec:supp_directional_matrices}

\fref{fig:supp_directional_matrices} reports every directional term that \textit{Selectivity} uses, and StyleComposer has positive target transfer on all three diagonal entries.
\begin{figure}[t]
\centering
\includegraphics[width=\linewidth]{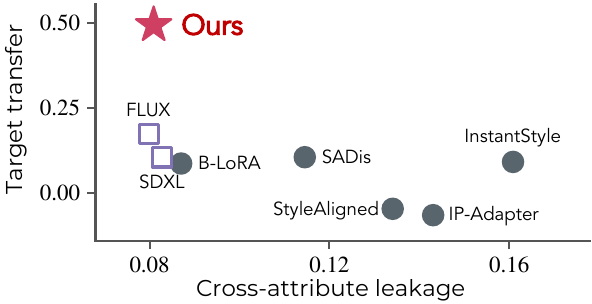}
\caption{\textbf{Target transfer and cross-attribute leakage.}
Each point averages the target transfer ($y$) and the non-target leakage ($x$) of one method over the three attributes, and the upper left is better.
Ours reaches the highest transfer and keeps its leakage as low as that of the prompt-only Flux baseline.}
\label{fig:transfer_leakage}
\end{figure}
\Cref{fig:transfer_leakage} averages the diagonal and off-diagonal entries of these matrices for each method.
Each image-conditioned baseline leaks more into the other attributes than prompt-only Flux and SDXL, and none of them moves the output toward the designated references as far as Flux does.
StyleComposer reaches about three times the transfer of Flux at the same leakage, so its gain does not come from injecting the references more strongly.

\subsection{Two-Reference Comparison}
\label{sec:supp_two_reference}
\begin{figure*}[t]
\centering
\includegraphics[width=\textwidth]{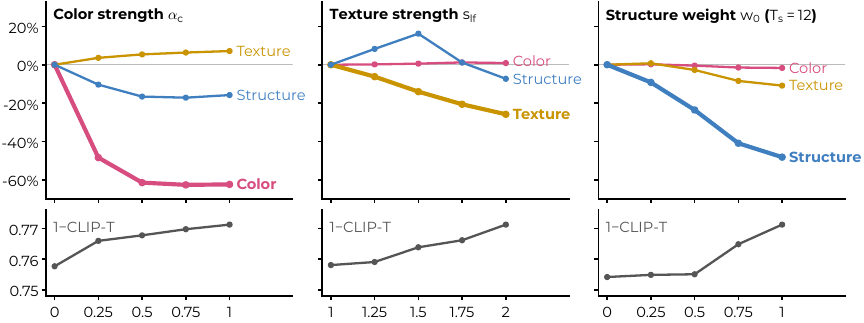}
\caption{\textbf{Response to each control.}
Each column raises one control from its lowest value to the benchmark setting while the other two stay at the benchmark setting, averaged over 40 cases.
The structure column varies the initial weight $w_0$ of the query blend and fixes the structure strength at $T_s=12$.
The top row shows the change of the color (MS-SWD), texture (grayscale CSD), and structure (depth) distances relative to the lowest setting, and the bottom row shows one minus CLIP-T.
The thick line marks the attribute that the control of each column targets, and it moves the most in every column.}
\label{fig:supp_strength_response}
\end{figure*}

The comparison in \cref{sec:two_reference} combines the five color references and the five texture references of the benchmark pools with four prompts, which gives 100 cases generated with seed 42.
The prompts describe a lighthouse on a snowy shore, a sailboat on the ocean, the New York City skyline, and a scenic waterfall.
Each texture reference has one medium descriptor, namely soft felt 3D rendering, rough oil pastel painting, layered paper-cut art, miniature diorama rendering, and felt-tip marker drawing, and every method receives the prompt followed by this descriptor.

StyleComposer keeps the setting of the main benchmark and generates at $512\times512$, and the other methods generate at $1024\times1024$.
SADis runs on SDXL~\cite{podell2024sdxl} with 50 sampling steps and guidance scale 5, and Dual-FDM runs on RealVisXL V5.0 with IP-Adapter~\cite{ye2023ip}, 30 steps, and guidance scale 5.
Both receive the color reference through their color input and the texture reference through their style input.
The multi-image generators take the color reference as the first image, the texture reference as the second, and the instruction ``Generate the following scene: [prompt] Use image 1 only for the color palette and image 2 only for texture and artistic style. Preserve the requested subject instead of copying objects from the reference images.''
OmniGen2 and Flux.2 (Klein) use 50 sampling steps and Qwen-Image-2.1 uses 40, with guidance scales of 5 for text and 2 for images in OmniGen2, 4 in Flux.2 (Klein), and 1 in Qwen-Image-2.1.

Color, texture, and prompt alignment are measured with MS-SWD, grayscale CSD, and CLIP-T, where CLIP-T always uses the prompt without the descriptor.
\textit{Composition} takes the harmonic mean of the three normalized similarities in each case and averages it over the 100 cases.
Each distance is normalized between its 5th and 95th percentiles over 500 outputs that StyleComposer, SADis, Dual-FDM, OmniGen2, and Flux.2 (Klein) generate for the same cases from the prompt alone, and every method shares these ranges.

\subsection{Additional Qualitative Comparisons}
\label{sec:supp_tier_qualitative}

\begin{figure*}[p]
\centering
\includegraphics[width=\textwidth]{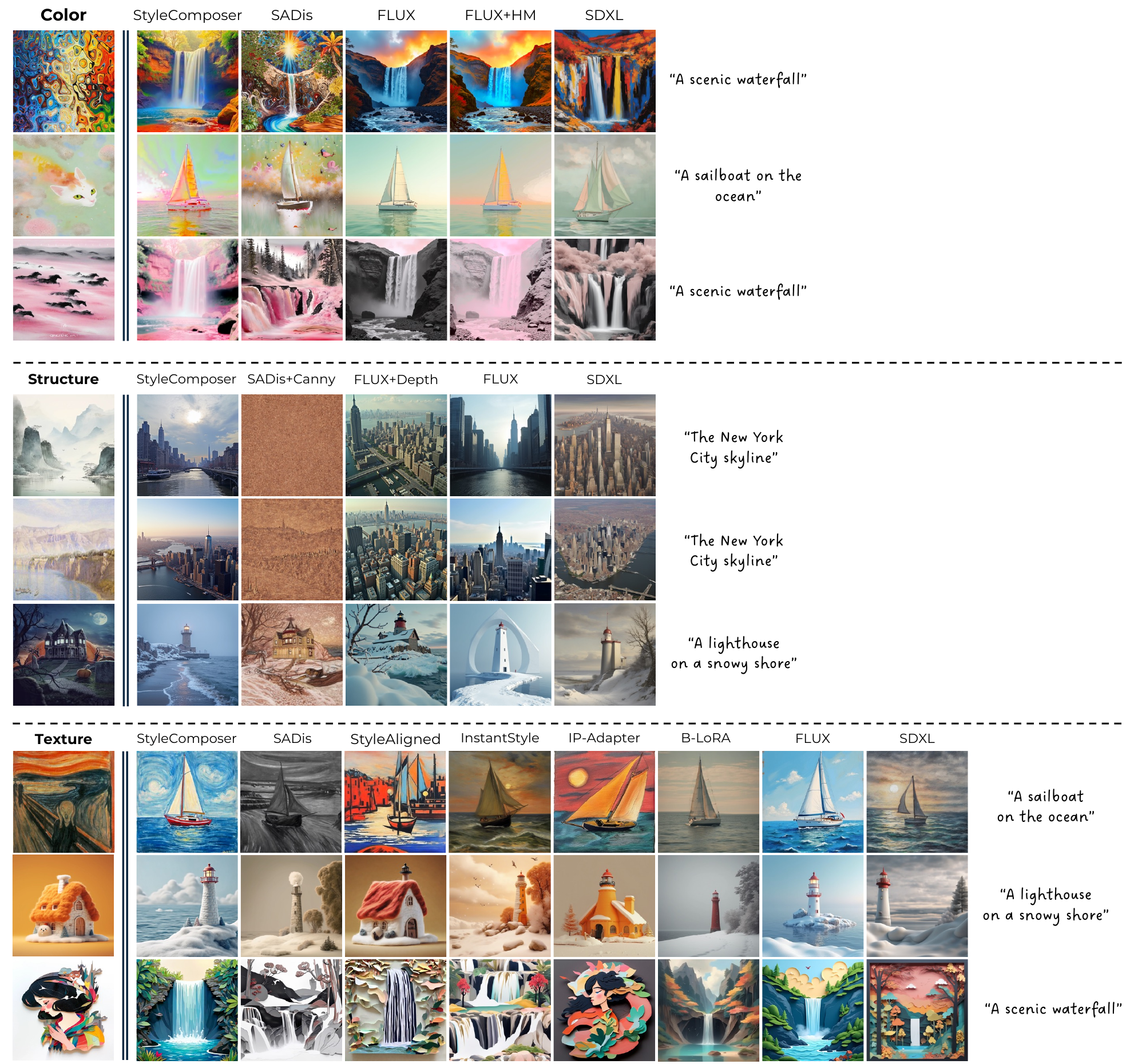}
\caption{\textbf{Single-reference comparisons for color, structure, and
texture.}
The leftmost column of each block is the designated reference, followed by
the methods supported in that tier; the prompt is shown at right. All
outputs in a row use the same prompt and seed. Flux+HM applies fixed
post-hoc CIE Lab histogram matching to the Flux output and serves as a
color-transfer reference point. For structure, SADis+Canny and Flux+Depth
augment their generators with image-based spatial control. Flux and SDXL receive textual descriptions of every requested attribute because they have no image input.}
\label{fig:supp_single_reference}
\end{figure*}

In the color-only rows of \fref{fig:supp_single_reference},
StyleComposer follows palettes that range from highly saturated to pastel
while retaining the prompt-specified scene. Post-hoc histogram matching can
also reproduce a palette, but it recolors a generation after its appearance
has already been formed. SADis sometimes brings in additional appearance from
the color reference, illustrating why palette similarity alone does not
measure selective color transfer. In the texture rows, several stylization baselines reproduce conspicuous surface characteristics, but they may also alter palette or scene appearance. StyleComposer transfers the rendering while preserving the requested content. In the structure rows, edge and depth controls impose strong spatial constraints, whereas the query route follows the coarse arrangement without copying the objects of the reference image.

\begin{figure*}[p]
\centering
\includegraphics[width=\textwidth]{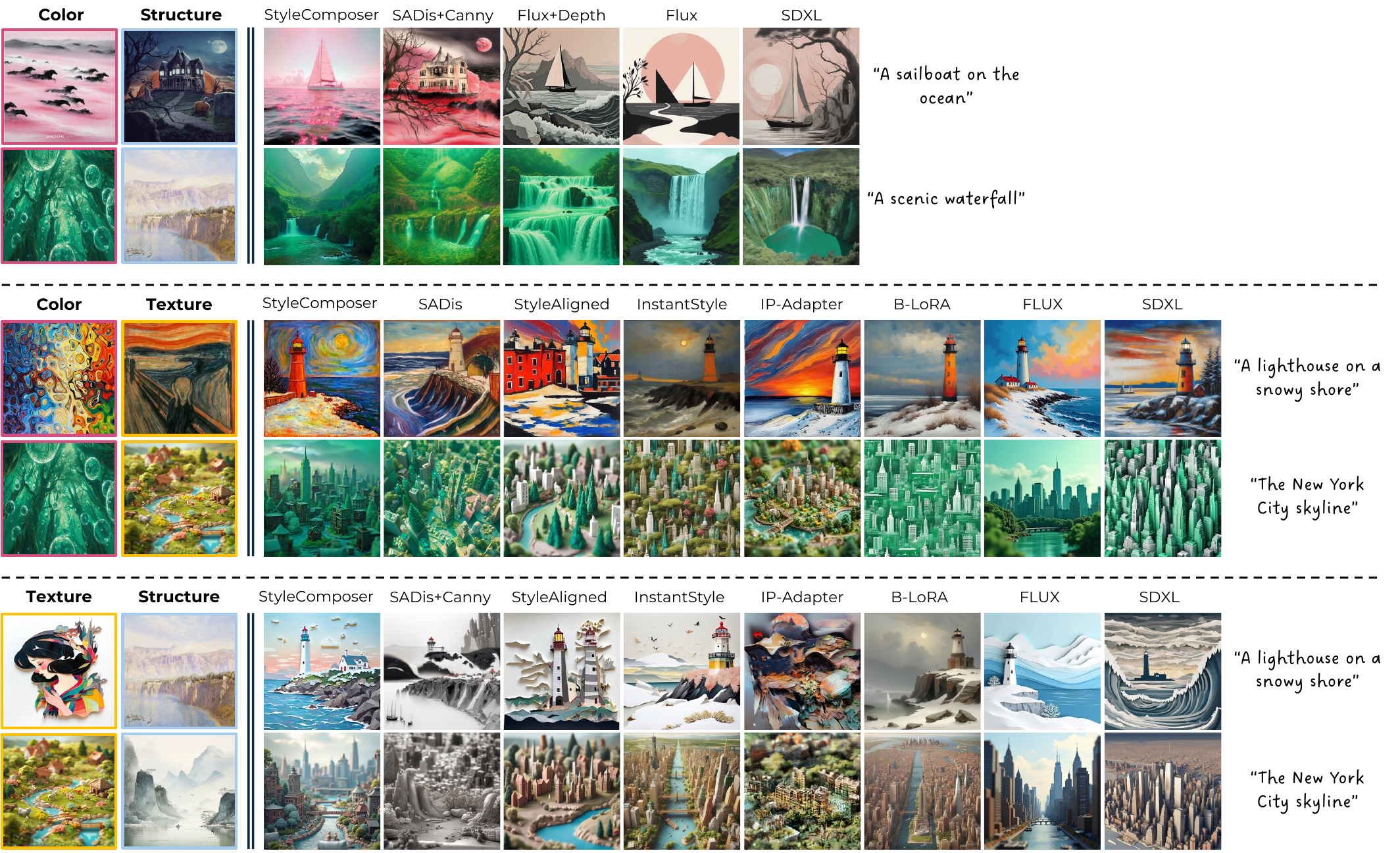}
\caption{\textbf{Dual-reference comparisons.}
The three blocks evaluate color+structure, color+texture, and
texture+structure composition. Reference columns are shown on the left and
the shared prompt on the right. Within each row, all methods use the same
prompt and seed and are evaluated against the same reference pair under the
capability-aware input mapping of \sref{sec:supp_input_mapping}.
SADis+Canny supplies structure through the official SDXL Canny ControlNet.
IP-Adapter uses the added depth ControlNet whenever structure is active,
including the texture+structure block. The Flux+Depth baseline in the color+structure block likewise conditions on the structure reference through depth ControlNet.}
\label{fig:supp_dual_reference}
\end{figure*}

\fref{fig:supp_dual_reference} makes the composition requirement more
apparent. In color+structure cases, separately adding Canny or depth control
can reproduce spatial cues but may introduce reference content or weaken the
requested palette. In color+texture cases, single-style methods frequently
favor either the rendering style or its associated colors. In
texture+structure cases, methods without a dedicated structure route often
preserve texture at the expense of layout, while stronger spatial controls
can suppress the intended rendering style. Across the three pair types,
StyleComposer more consistently retains the prompt object while combining
the two designated attributes. These examples complement the results by reference combination in \cref{tab:supp_tier_composition}.

\begin{figure*}[p]
\centering
\includegraphics[width=\textwidth]{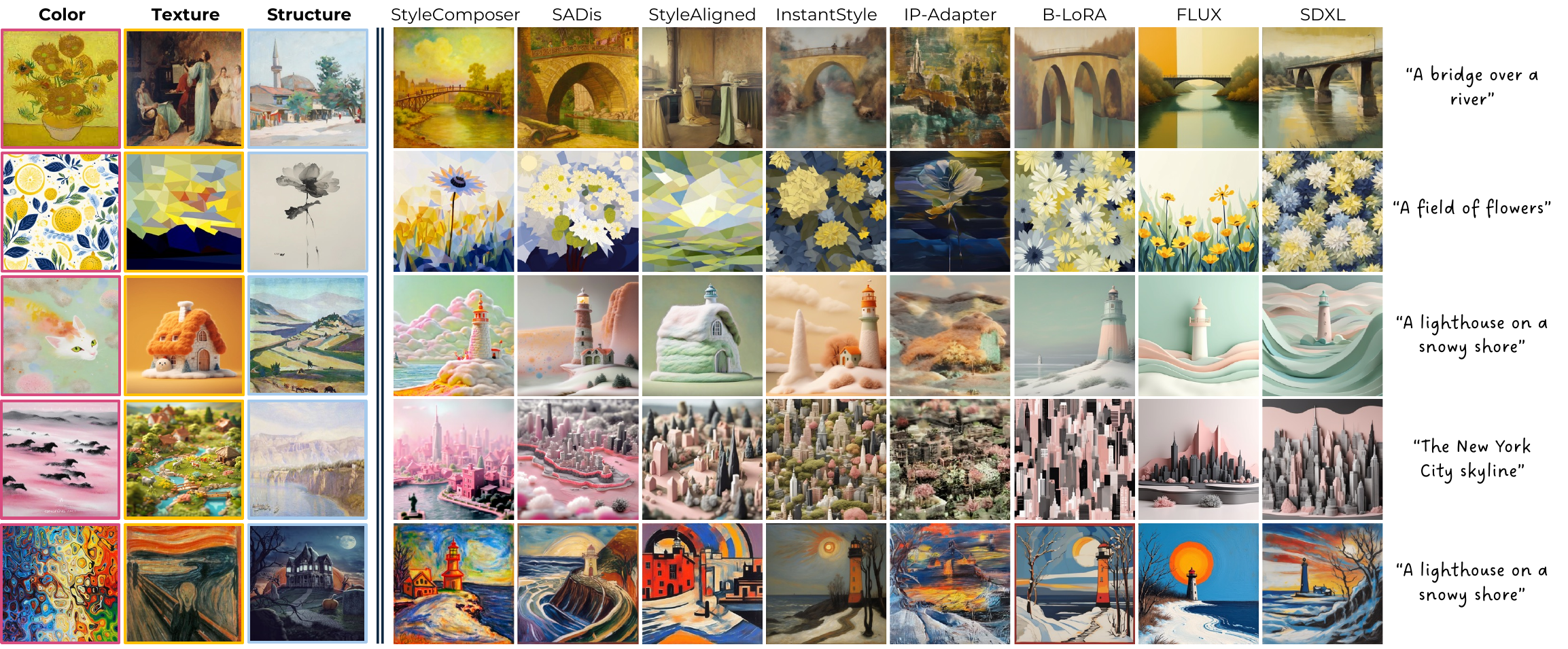}
\caption{\textbf{Additional three-reference comparisons.}
Each row provides separate color, texture, and structure references followed
by outputs from all eight methods. The prompt is shown at right. Within each
row, all methods use the same references, prompt, and seed under the input
mapping in \sref{sec:supp_input_mapping}; IP-Adapter is augmented with
depth ControlNet for the structure reference.}
\label{fig:supp_triple_reference}
\end{figure*}

\fref{fig:supp_triple_reference} extends the comparison to five more three-reference cases. Baselines often reproduce one conspicuous
cue while coupling it to the reference's remaining appearance or weakening
another requested cue. StyleComposer combines the designated palette, rendering style, and coarse layout more consistently than the baselines and keeps the subject named in the prompt. 

\section{Additional Analyses}
\label{sec:supp_additional_results}

\subsection{Reference Replacement}
\label{sec:supp_replacement}
\begin{table*}[t]
\centering
\footnotesize
\setlength{\tabcolsep}{8pt}
\caption{\textbf{Within-method comparison of image conditioning and textual substitution.} Descriptions of the same references replace the color image route, the structure image route, or both, while the texture image route and the texture-medium descriptor stay fixed.}
\label{tab:supp_text_substitution}
\begin{tabular*}{\textwidth}{@{\extracolsep{\fill}}lccccccc@{}}
\toprule
Variant & Color source & Structure source
& \textit{Composition}$\uparrow$ & \textit{Selectivity}$\uparrow$
& MS-SWD$\downarrow$ & gCSD$\downarrow$ & Depth$\downarrow$ \\
\midrule
Full (StyleComposer) & Image & Image
& \textbf{0.621} & \textbf{0.745}
& \textbf{5.32} & 0.439 & \textbf{0.088} \\
Color to text & Text & Image
& 0.395 & 0.607
& 11.15 & \textbf{0.367} & 0.105 \\
Structure to text & Image & Text
& 0.568 & 0.512
& 5.56 & 0.444 & 0.139 \\
Color and structure to text & Text & Text
& 0.449 & 0.486
& 9.51 & 0.417 & 0.144 \\
\bottomrule
\end{tabular*}
\end{table*}
\begin{table*}[!t]
\centering
\footnotesize
\setlength{\tabcolsep}{6pt}
\caption{\textbf{Detailed design ablation} on the same 54 three-reference
cases as the main paper. All variants use the final prompt protocol and the
normalization fixed from the main benchmark. Attribute columns report the
primary distance for each axis.}
\label{tab:supp_design_ablation}
\begin{tabular*}{\textwidth}{@{\extracolsep{\fill}}lcccccc@{}}
\toprule
Variant
& \textit{Composition}$\uparrow$
& \textit{Selectivity}$\uparrow$
& MS-SWD$\downarrow$
& gCSD$\downarrow$
& Depth$\downarrow$
& 1$-$CLIP-T$\downarrow$ \\
\midrule
Single K/V route
& 0.175 & 0.371 & 15.509 & 0.608 & \underline{0.116} & \textbf{0.749} \\
Overlapped route windows
& 0.150 & \textbf{0.611} & 5.926 & 0.527 & \textbf{0.100} & 0.848 \\
Full-band texture K
& 0.061 & 0.357 & \textbf{5.776} & 0.489 & 0.718 & 0.881 \\
Q/K AdaIN
& \underline{0.351} & 0.508 & \underline{5.819} & \underline{0.427} & 0.191 & 0.819 \\
Without texture-attention cap
& 0.325 & 0.515 & 5.955 & \textbf{0.402} & 0.208 & 0.823 \\
\midrule
Full (StyleComposer)
& \textbf{0.583} & \underline{0.585} & 5.887 & 0.442 & 0.119 & \underline{0.788} \\
\bottomrule
\end{tabular*}
\end{table*}

The replacement test of \cref{sec:replacement} uses the four prompts of \cref{sec:supp_two_reference}, five three-reference combinations for each prompt, and the seeds 42 and 1234.
For each combination, one other color, texture, and structure reference from the same pools forms the replaced condition, and the remaining inputs and the default strengths stay fixed.
Within each prompt, a fixed cyclic assignment uses every reference of each pool once as the original and once as the replacement.
StyleComposer generates all 160 outputs at $512\times512$ with 28 sampling steps and guidance scale 3.5, and the prompts contain no medium descriptor.
The color distance is the sliced Wasserstein distance between the Lab colors of two images, and the texture distance is the cosine distance between their grayscale CSD embeddings.
The structure distance is one minus the correlation between their depth maps, each centered at its mean.
Each difference in \cref{eq:response} compares the two outputs against the same reference, so an offset that a distance gives to one reference cancels.
The two differences also take the outputs in opposite order, so an offset that a distance gives to one output cancels in their sum.

\subsection{Per-Attribute Strength}
\label{sec:supp_strength}

We measure how far each control moves the output on 20 reference--prompt combinations with two seeds each.
The combinations pair ten prompts, such as a steam train crossing a stone bridge, with references from the benchmark pools, and StyleComposer generates every output at $1024\times1024$ with 28 sampling steps.
For each of the 40 cases, the references, the prompt, the medium descriptor, and the seed stay fixed, and one control takes five values from its lowest setting to the benchmark setting while the other two stay at the benchmark setting.
The color strength $\alpha_c$ ranges from 0 to 1, and the texture strength $s_{\mathrm{lf}}$ ranges from 1 to 2.
The structure route has two variables.
The structure strength $T_s$ sets how many early steps receive the reference queries, and the initial weight $w_0$ scales the cosine weight of \cref{eq:supp_structure_decay}, so it sets how strongly the reference queries replace the generation queries at the first step before the weight decays.
This test fixes $T_s=12$ and raises $w_0$ from 0, which removes the structure route as $T_s=0$ does, to the benchmark value of 1, and \cref{tab:supp_route_settings} varies $T_s$ with $w_0=1$.

Each control mainly changes the attribute it targets.
Between the lowest and the benchmark setting, each control lowers the distance of its own attribute by 62\% for color, 26\% for texture, and 48\% for structure, and this distance falls in 39, 39, and 33 of the 40 cases (\cref{fig:supp_strength_response}).
Over the same range, the other two distances change by at most 16\%.
CLIP-T at the benchmark setting is at most 0.017 below its value at the lowest setting.

\begin{figure*}[p]
\centering
\includegraphics[width=\textwidth]{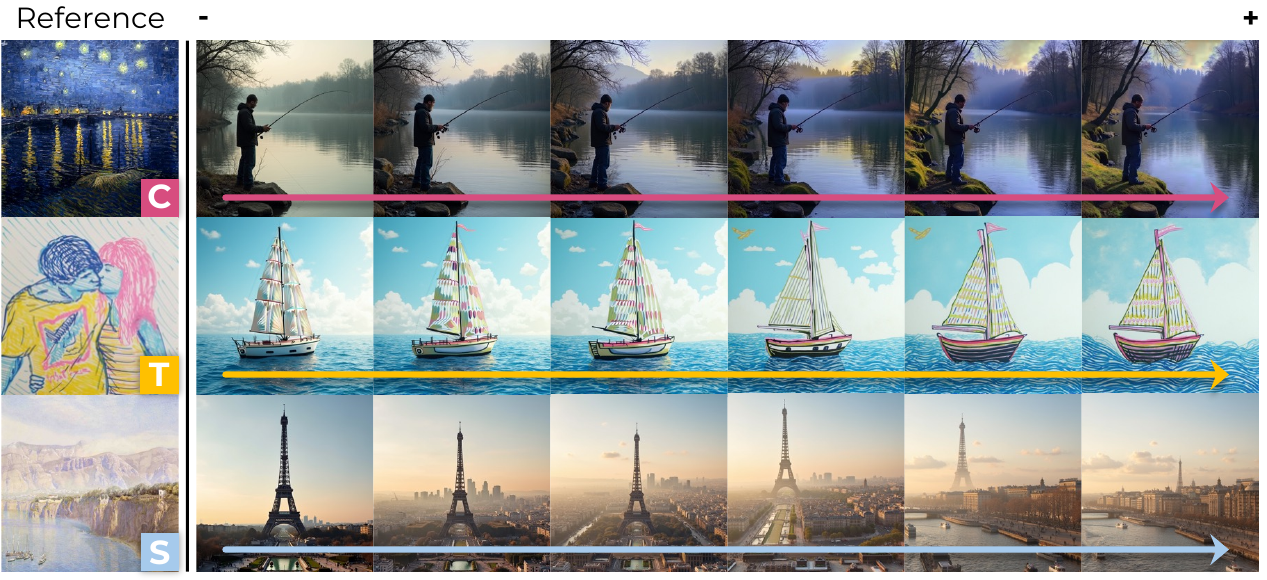}
\caption{\textbf{Strength of one attribute.}
Each row raises the strength of the attribute marked on its reference over six settings, with the references, the prompt, and the seed fixed, and C, T, and S mark the color, texture, and structure references.
As the strength rises, the output follows that reference more closely on the marked attribute.}
\label{fig:supp_strength_single}

\vspace{2em}
\includegraphics[width=\textwidth]{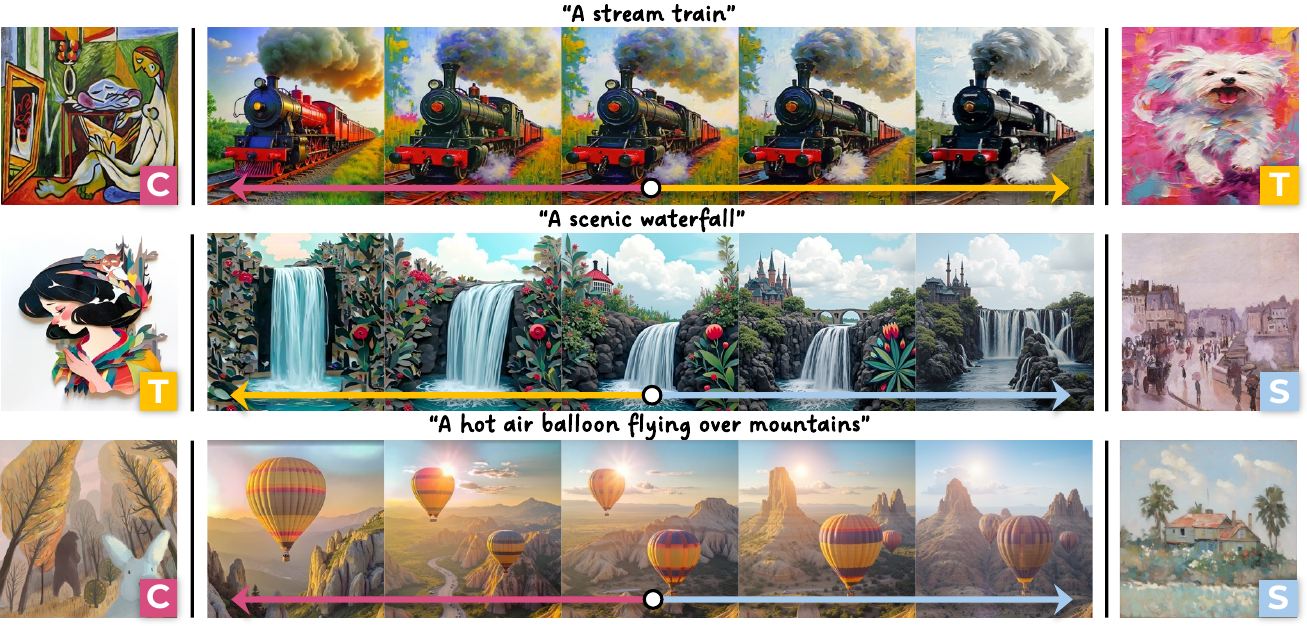}
\caption{\textbf{Strengths of two attributes in opposite directions.}
Each row fixes the prompt above it, the seed, and the two references at its sides, and C, T, and S mark the color, texture, and structure references.
From left to right, the strength of the left reference decreases and that of the right reference increases, so the output moves from the attribute of the left reference to that of the right.}
\label{fig:supp_strength_pairwise}
\end{figure*}

\Cref{fig:supp_strength_single,fig:supp_strength_pairwise} show the complete sweeps from which \cref{fig:control} takes three settings per row.
In each row, the references, the prompt, and the seed stay fixed, and only the strengths change.
As the color strength rises in \cref{fig:supp_strength_single}, the lake scene goes from pale, hazy tones to the deep blues of the color reference.
As the texture strength rises, the sailboat turns from a smooth rendering into the hand-drawn lines of the texture reference, and as the structure strength rises, the Eiffel Tower recedes toward the horizon while a river crosses the lower part of the frame, as in the structure reference.
In \cref{fig:supp_strength_pairwise}, each step lowers the strength of one reference and raises that of the other, so the output moves gradually from one attribute emphasis to the other.

\subsection{Ablations}
\label{sec:supp_ablations}
\begin{figure*}[t]
\centering
\includegraphics[width=\textwidth]{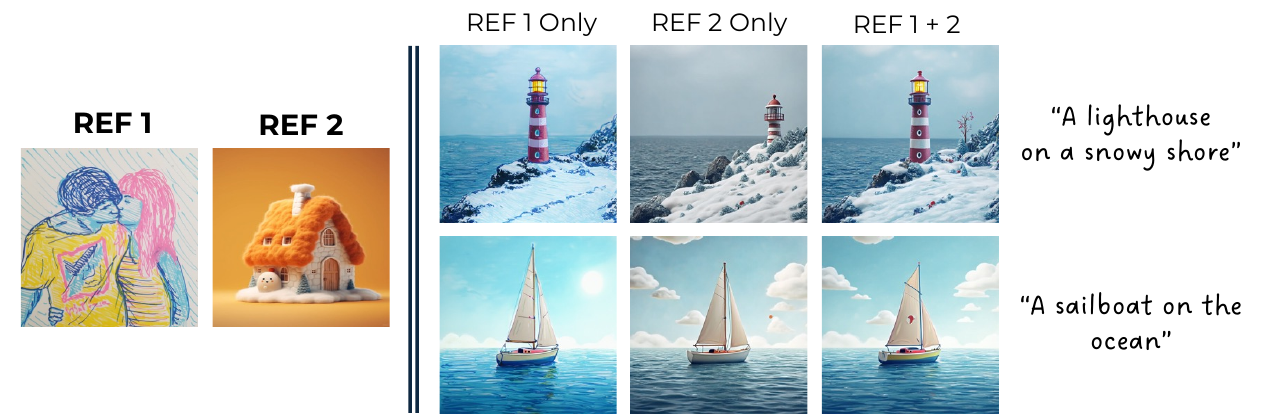}
\caption{\textbf{Two texture references in one route.}
Reference 1 has a line-based marker appearance and reference 2 a soft, yarn-like surface.
Mixed through the texture route, the upper example follows reference 2 more closely and the lower example balances the two.
A route for each finer element would set this balance directly, which we leave to future work.}
\label{fig:supp_limitations}
\end{figure*}

\noindent\textbf{Route removal.}\label{sec:supp_route_contribution}
The route-removal experiment in the main paper evaluates each pathway on the
same 54 three-reference cases used for the design ablation below. Removing
the color route disables latent chromatic alignment, removing the texture
route disables reference K/V routing, and removing the structure route
disables reference-Q guidance. The texture-medium descriptor is retained
when K/V routing is removed, so the comparison isolates the contribution of
the texture image pathway beyond its short textual category cue. 

\smallskip
\noindent\textbf{Prompt descriptions in place of routes.}\label{sec:supp_text_substitution}
We test whether color and structure images can be replaced by descriptions
of the same references. Starting from the final StyleComposer configuration,
we disable the corresponding image route and append either the palette
description $d_c$, the layout description $d_s$, or both to the prompt.
The texture image route and texture-medium descriptor remain unchanged in
every variant. We use the descriptions given to the baselines, keep all
references, content prompts, seeds, and remaining routes fixed, and evaluate
the removed attribute against its original image reference rather than
against the description. All variants are evaluated on the complete set of
250 three-reference cases without per-case tuning.

\begin{figure*}[p]
\centering
\includegraphics[width=0.85\textwidth]{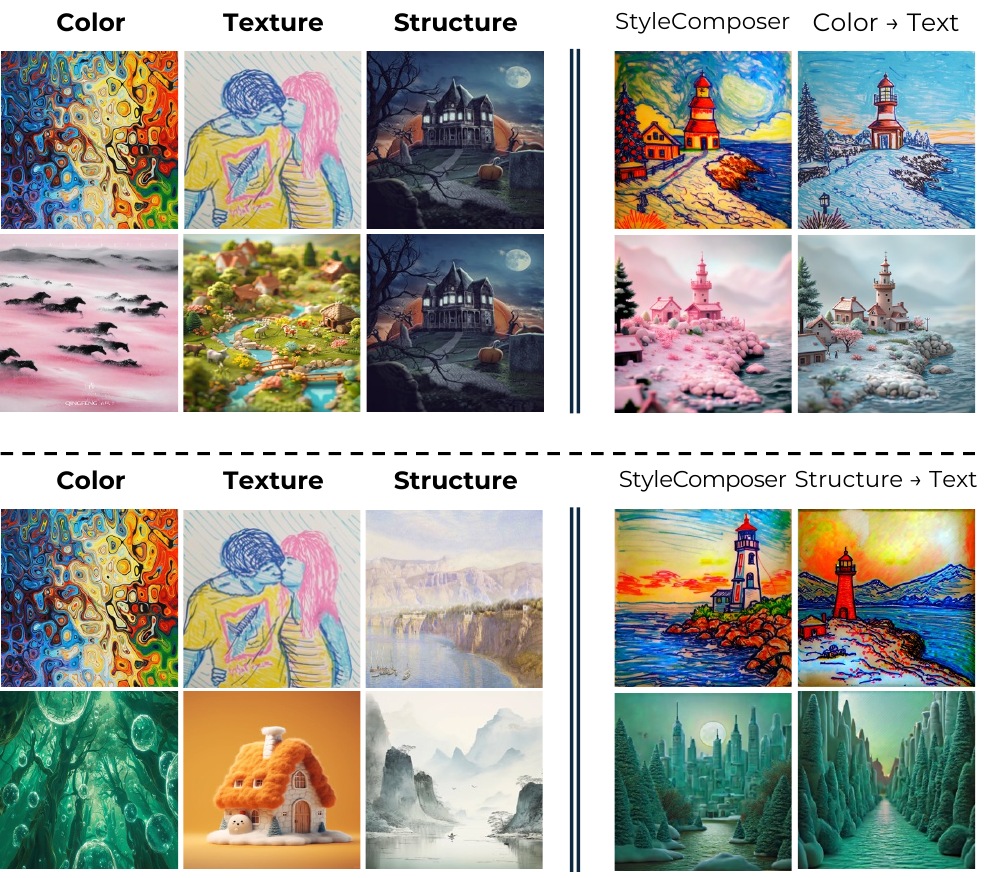}
\caption{\textbf{Image routing versus textual attribute substitution.}
Each row keeps the reference triplet, content prompt, seed, and remaining
image routes fixed. In the upper block, \emph{Color to Text} replaces the
color route with a palette description derived from the same reference. In
the lower block, \emph{Structure to Text} replaces the structure route with a
layout description. Text conveys coarse attribute cues, whereas the full
model more closely follows the reference-specific palette or spatial
arrangement.}
\label{fig:supp_text_substitution}
\end{figure*}

\tref{tab:supp_text_substitution} quantifies the comparison, and
\fref{fig:supp_text_substitution} provides paired qualitative examples.
Replacing either image route weakens fidelity to its designated reference.
Color-to-text more than doubles MS-SWD, while structure-to-text increases
depth-layout distance despite retaining the corresponding layout
description. 
Replacing both routes produces the interface used by a
single-style-image method, with texture supplied by an image and color and
structure supplied through text, but it remains below the full model in
\textit{Composition} and \textit{Selectivity}. Text descriptions can convey coarse palettes and layouts, yet they do not replace the image routes.

\smallskip
\noindent\textbf{Design ablation.}\label{sec:supp_design_ablation}
The single-route variant sends all attributes through one low-frequency K/V pathway. Overlapped windows
retain the three routes but activate them throughout denoising, removing
their temporal coordination. Full-band texture K keeps all positional frequencies, Q/K AdaIN aligns the query statistics as well as the key statistics, and the final variant removes the step-dependent cap on
attention to texture-reference tokens. Each alternative reduces
\textit{Composition} for a different reason. The single K/V route loses color
fidelity. Full-band K severely disrupts structure and prompt alignment,
and Q/K AdaIN and uncapped texture attention also weaken both. Overlapped windows attain higher
\textit{Selectivity} and lower depth distance, but substantially reduce prompt
alignment and joint \textit{Composition}. 

\subsection{Other Backbones}
\label{sec:supp_backbones}

\begin{figure*}[p]
\centering
\footnotesize
\setlength{\tilew}{\dimexpr((\textwidth-12pt)/2-14pt)/3\relax}%
\setlength{\tabcolsep}{0pt}%
\newcommand{\bb}[1]{\includegraphics[width=\tilew]{figures/newexp/bbf_#1.jpg}}%
\newcommand{\bbl}[1]{\rotatebox{90}{\makebox[\tilew]{#1}}}%
\newcommand{\bbblock}[2]{%
\begin{tabular}[t]{@{}p{10pt}@{\hspace{1pt}}c@{\hspace{1pt}}c@{\hspace{1pt}}c@{}}
& \bb{#1_cref} & \bb{#1_tref} & \parbox[b]{\tilew}{\raggedright Color and texture references for ``#2''}\\[1pt]
& No reference & + Texture & + Color\\[1pt]
\bbl{Qwen-Image-2.1} & \bb{#1_qwen_basetext} & \bb{#1_qwen_t} & \bb{#1_qwen_ct}\\[-1pt]
\bbl{Z-Image} & \bb{#1_zimage_basetext} & \bb{#1_zimage_t} & \bb{#1_zimage_ct}\\[-1pt]
\bbl{Flux.2 (Klein)} & \bb{#1_flux2_basetext} & \bb{#1_flux2_t} & \bb{#1_flux2_ct}
\end{tabular}}%
\bbblock{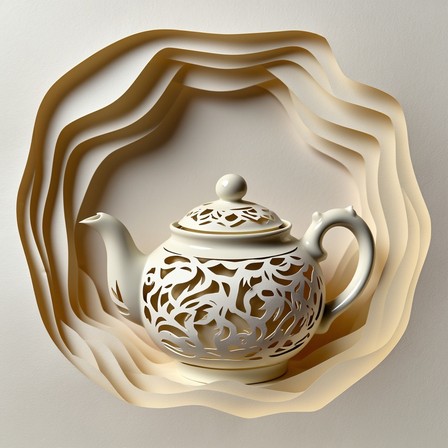}{A ceramic teapot, layered paper-cut art}\hspace{12pt}\bbblock{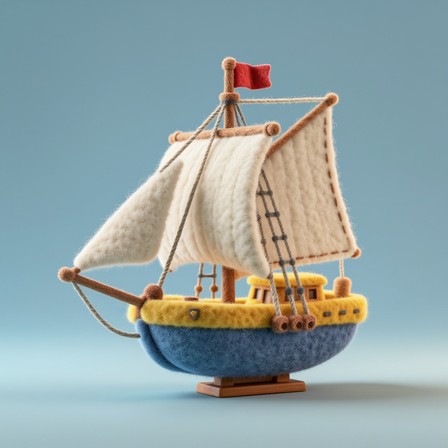}{A sailing ship, soft felt 3D rendering}
\caption{\textbf{Color and texture routes on other backbones.} Each block fixes one prompt with the medium descriptor of the texture reference, and each row is a backbone. No reference uses the prompt alone, + Texture adds the texture route, and + Color adds the color route to it. The texture route adds paper layers and a felt surface beyond what the descriptor alone gives, and the color route then moves the palette to that of the color reference on every backbone.}
\label{fig:supp_backbones}
\end{figure*}

The color route needs only a VAE latent and its chromatic subspace, and the texture route needs only joint attention over image tokens with rotary positions.
We apply these two routes to three other DiT backbones, Qwen-Image-2.1~\cite{qwenimage2026}, Z-Image~\cite{zimage2025}, and the 4B base model of Flux.2 (Klein)~\cite{flux2klein2026}, each with its native sampler.
For each VAE, we recompute the chromatic subspace from 512 uniform-color images as the top three principal directions of their mean latents.
Each route keeps the fraction of the sampling steps it covers in the 28 steps of Flux.1-dev, so the texture route starts after 3/28 of the steps and the color route after 16/28.
On Qwen-Image-2.1, the references enter only through the routes, and its native image conditioning is off.
We fix the route settings of each backbone once and use them for every case.
Each backbone generates at $1024\times1024$ with seed 42, using 40 sampling steps for Qwen-Image-2.1 and Z-Image and 50 for Flux.2 (Klein).
Every column of \cref{fig:supp_backbones} uses the prompt with the medium descriptor of the texture reference, so the columns differ only in which routes are active.
Beyond what the descriptor gives, the texture route adds the paper layers of the cut-paper portrait and the felt surface of the felt house, most visibly on Z-Image and Flux.2 (Klein).
On all three backbones, adding the color route then moves the palette toward the pastel tones of the cat painting and toward the greens of the forest painting.

\subsection{Two References in One Route}
\label{sec:supp_two_in_one}

The texture route can also take two references at once.
In \cref{fig:supp_limitations}, mixing a line-based marker drawing with a soft, yarn-like surface gives outputs that combine the two, and the balance between them differs between the two examples.


\end{document}